\documentclass{article}

\usepackage{microtype}
\usepackage{graphicx}
\usepackage{booktabs} 

\usepackage{hyperref}

\usepackage[accepted]{icml2024}

\usepackage{amsmath}
\usepackage{amssymb}
\usepackage{mathtools}
\usepackage{amsthm}
\usepackage{microtype}
\usepackage{graphicx}
\usepackage{booktabs} 
\usepackage{hyperref}       
\usepackage{url}            
\usepackage{booktabs}       
\usepackage{amsfonts}       
\usepackage{nicefrac}       
\usepackage{microtype}      
\usepackage{xcolor}         
\usepackage{enumitem}
\usepackage{amsmath}
\usepackage{amsthm}
\usepackage{wrapfig}
\usepackage{caption}
\usepackage{subcaption}

\newtheorem{Theorem}{Theorem}

\newtheorem{Definition}{Definition}

\usepackage{multicol}
\usepackage{multirow}
\usepackage{graphicx}
\usepackage{amsfonts}
\usepackage{latexsym}
\usepackage{algorithm}
\usepackage{multirow,tikz,amsmath,comment}
\usepackage{enumitem}
\usepackage[utf8]{inputenc} 
\usepackage[T1]{fontenc}    
\usepackage{hyperref}       
\usepackage{url}            
\usepackage{booktabs}       
\usepackage{amsfonts}       
\usepackage{nicefrac}       
\usepackage{microtype}      
\usepackage{sidecap}
\usepackage{natbib}
\usepackage{mathtools}
\usepackage{txfonts}
\usepackage{multicol}
\usepackage{arydshln}
\newcommand\independent{\protect\mathpalette{\protect\independenT}{\perp}}
\def\independenT#1#2{\mathrel{\rlap{$#1#2$}\mkern2mu{#1#2}}}

\usepackage{hyperref}

\usepackage{amsmath}
\usepackage{amssymb}
\usepackage{mathtools}
\usepackage{amsthm}

\newcommand{\Biwei}[1]{{ \color{blue} Biwei: #1}}
\usepackage[capitalize,noabbrev]{cleveref}

\theoremstyle{plain}

\theoremstyle{definition}

\theoremstyle{remark}

\usepackage[textsize=tiny]{todonotes}

\icmltitlerunning{Advancing Counterfactual Inference through Nonlinear Quantile Regression}

\begin{document}

\twocolumn[
\icmltitle{Advancing Counterfactual Inference through Nonlinear Quantile Regression}



\icmlsetsymbol{equal}{*}

\begin{icmlauthorlist}
\icmlauthor{Shaoan Xie}{equal,cmu}
\icmlauthor{Biwei Huang}{equal,ucsd}
\icmlauthor{Bin Gu}{mbz}
\icmlauthor{Tongliang Liu}{syn,mbz}
\icmlauthor{Kun Zhang}{cmu,mbz}

\end{icmlauthorlist}

\icmlaffiliation{cmu}{Carnegie Mellon University}
\icmlaffiliation{syn}{The University of Sydney}
\icmlaffiliation{ucsd}{University of California San Diego}
\icmlaffiliation{mbz}{Mohamed bin Zayed University of Artificial Intelligence}

\icmlcorrespondingauthor{Kun Zhang}{kunz1@cmu.edu}

\icmlkeywords{Machine Learning, ICML}

\vskip 0.3in
]



\printAffiliationsAndNotice{\icmlEqualContribution} 

\begin{abstract}
The capacity to address counterfactual "what if" inquiries is crucial for understanding and making use of causal influences. Traditional counterfactual inference, under Pearls' counterfactual framework, typically depends on having access to or estimating a structural causal model. Yet, in practice, this causal model is often unknown and might be challenging to identify. Hence, this paper aims to perform reliable counterfactual inference based solely on observational data and the (learned) qualitative causal structure, without necessitating a predefined causal model or even direct estimations of conditional distributions. To this end, we establish a novel connection between counterfactual inference and quantile regression and show that counterfactual inference can be reframed as an extended quantile regression problem. Building on this insight, we propose a practical framework for efficient and effective counterfactual inference implemented with neural networks under a bi-level optimization scheme. The proposed approach enhances the capacity to generalize estimated counterfactual outcomes to unseen data, thereby providing an upper bound on the generalization error. Furthermore, empirical evidence demonstrates its superior statistical efficiency in comparison to existing methods. 
Empirical results conducted on multiple datasets offer compelling support for our theoretical assertions.
\end{abstract}

\section{Introduction}

Understanding and making use of cause-and-effect relationships play a central role in scientific research, policy analysis, and everyday decision-making. Pearl's causal ladder \citep{Pearl2000} delineates the hierarchy of prediction, intervention, and counterfactuals, reflecting their increasing complexity and difficulty. Counterfactual inference, the most challenging level, allows us to explore what would have happened if certain actions or conditions had been set to a different value, providing valuable insights into the underlying causal relationships between variables.

\begin{figure}%
    \centering
{\includegraphics[scale=0.3]{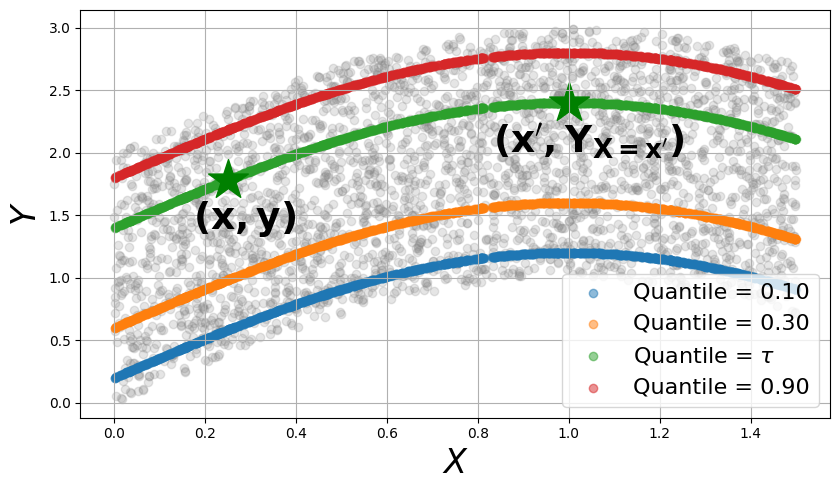} }%
\setlength{\abovecaptionskip}{0pt}    \caption{\small Illustration of our proposed quantile-based counterfactual estimation ($Z$ is omitted for illustration purpose). For a sample of interest ($X=x,Z=z,Y=y$), we estimate the quantile $\tau=P(Y\leq y|X=x,Z=z)=0.70$ with factual observations. Then the counterfactual outcome $Y_{X=x^\prime}$ is equal to the value $y^\prime$ which satisfy $P(Y\leq y^\prime|X=x^\prime, Z=z)=\tau$.
    }
    \vspace{-0.45cm}
    \label{fig:illus}%
\end{figure}

Conventional approaches to counterfactual inference often rely on having access to or estimating a structural causal model (SCM). For instance, within Pearls' framework \cite{Pearl2000}, if the SCM is given, the typical three-step procedure involves first estimating the noise value of the sample of interest, then modifying the model, and finally computing the counterfactual value using the estimated noise within the adjusted SCM. 
Unfortunately, the structure causal model is often unavailable. To tackle this issue, various approaches have been proposed for estimating the SCM using observational data \citep{pawlowski2020deep,de2022deep,sanchez2022diffusion,song2020denoising,chao2023interventional,lu2020sample,nasr2023counterfactual,khemakhem2021causal,xia2022neural,de2021transport,melnychuk2022causal}. However, general causal models without specific functional constraints may lack identifiability \citep{zhang2015estimation}. In other words, there is no guarantee that the learned model is the true SCM, making the counterfactual predictions unreliable. Moreover, current methods often rely on specific distributional assumptions regarding the noise terms, thereby restricting the model class.

This paper aims to tackle the challenge of reliable counterfactual inference without relying on a predefined structural causal model or even direct estimations of conditional distributions. To this end, we establish a novel connection between counterfactual inference and quantile regression. 
Specifically,
suppose there are three random variables $X,Y,Z$ and a latent noise variable $E_{Y}$, where $X,Z,E_{Y}$ cause the outcome $Y$, and suppose that the observed evidence is $\langle X=x, Y=y, Z=z \rangle$. We find that the counterfactual outcome of $Y$ if $X$ had been intervened to $x'$ given the observed evidence, denoted by $Y_{X=x'} | X=x, Y=y, Z=z$, is equal to the $\tau$-th quantile of the conditional distribution $P(Y|X=x', Z=z)$, where $Y=y$ is the $\tau$-th quantile of $P(Y|X=x, Z=z)$, i.e.,
$\tau=P(Y\leq y |X=x, Z=z)$, under mild conditions. Consequently, we reframe counterfactual inference as an extended quantile regression problem, which yields improved statistical efficiency compared to existing methods. An illustration is provided in Fig. \ref{fig:illus}. Furthermore, our approach enhances the generalization ability of estimated counterfactual outcomes for unseen data and provides an upper bound on the generalization error. Our contributions can be summarized as follows.
\begin{itemize}[leftmargin=15pt,topsep=-2pt,itemsep=-2pt]
    \item  We introduce a novel framework that formulates counterfactual inference as an extended quantile regression problem. This framework is implemented using neural networks within a bi-level optimization scheme, offering superior statistical efficiency. 
    \item  
    We assess the generalization capacity of our proposed approach to previously unseen data and establish an upper bound on the generalization error.
    \item  We conduct extensive experiments to validate our theories and showcase the effectiveness of our proposed method in diverse scenarios.
\end{itemize}



\vspace{-1mm}
\section{Problem Formulation and Related Work} 
\vspace{-1.5mm}

In this section, we introduce key concepts relevant to our study, including Pearl's three-step procedure for counterfactual inference, the technique of quantile regression, and recent works in counterfactual inference under Pearl's procedure. Below, we first give a formal definition of counterfactual outcomes.

\begin{Definition}[Counterfactual outcomes \citep{Pearl2000}]
    Suppose $X$, $Y$, and $Z$ are random variables, where $X$ causes $Y$, and $Z$ is a set of common causes to $X$ and $Y$. Given observations $\langle X=x, Y=y, Z=z \rangle$, the counterfactual outcome of $Y$ is defined as the value of $Y$ if $X$ had been set to a different value $x'$ and is mathematically represented as $Y_{X=x'}|Y=y, X=x, Z=z$. 
\end{Definition}

\paragraph{Pearl's Three-Step Procedure for Counterfactual Inference}

In the context of a SCM, \citet{Pearl2000, Pearl2016Primer} introduced a three-step procedure to address such counterfactual reasoning.

Suppose the SCMs $Y = f_Y(X, Z,E_Y)$, $X = f_X(Z, E_X)$, and $Z = E_Z$ are given, denoted by $M$, and that we have evidence $\langle X=x, Y=y, Z=z \rangle$. The following steps outline the process of counterfactually inferring $Y$ if we had set $X=x'$ \citep{Pearl2000, Pearl2016Primer}:
\begin{itemize}[leftmargin=15pt,topsep=-2pt,itemsep=-1pt]
	\item Step 1 (abduction): Use the evidence $\langle X=x, Y=y, Z=z \rangle$ to determine the value of the noise $E_Y=e$.
	\item Step 2 (action): Modify the model, $M$, by removing the structural equations for the variables in $X$ and replacing them with the functions $X=x'$, thereby obtaining the modified model, $M_{x'}$.
	\item Step 3 (prediction): Use the modified model, $M_{x'}$, and the estimated noise $e$ to compute the counterfactual of $Y$ as $Y_{X=x^\prime}=f_Y(x^\prime, z, e)$.
\end{itemize}

Note that Step 1  performs deterministic counterfactual reasoning, focusing on counterfactuals of a single unit of the population, where the value of $E_{Y}$ is determined.

\vspace{-2mm}
\paragraph{Deep Counterfactual Inference } Various deep-learning approaches have been proposed for estimating the SCM and noise values using observational data and accordingly perform the three-step procedure for counterfactual inference. 
\cite{khemakhem2021causal} proposes to use autoregressive flow to perform causal discovery by comparing the likelihood and infer the noise for counterfactual inference by inverting the flow. \cite{javaloy2023causal} presents a mechanism to embed additional causal knowledge when learning the causal autoregressive flow. 
CFQP \citep{de2022deep} considers a setting when the background variables are categorical and employs the Expectation-Maximization framework to predict the cluster of the sample and perform counterfactual inference with the regression model trained on the specific cluster. CTRL \citep{lu2020sample} and BGM \citep{nasr2023counterfactual,nasr2023counterfactual2} show that the counterfactual outcome is identifiable when the SCM is monotonic w.r.t. the noise term. In particular, BGM uses conditional spline flow to mimic the generation process and performs counterfactual inference by reversing the flow. 
DeepSCM \cite{pawlowski2020deep} proposes to use variational inference and normalizing flow to infer the noise variable. DiffSCM \cite{sanchez2022diffusion} proposes to match the observation distribution with a conditional diffusion model and infer the noise by reversing the diffusion process, but it only allows one cause.  DCM\cite{chao2023interventional} generalizes DiffSCM to support multiple causes of a single variable.
G-NCM \cite{xia2022neural} extends the neural causal model \cite{xia2021causal}
to estimate the counterfactual distribution. \cite{melnychuk2022causal} proposes a counterfactual domain
confusion loss to address confounding bias and uses a transformer to perform counterfactual inference for long-range time-series data. \cite{ribeiro2023high} proposes a hierarchical latent mediator model for counterfactual image generation task. Individual treatment effect (ITE) \citep{johansson2016learning,yoon2018ganite,bica2020estimating,SITE2018,li2022contrastive,lu2020reconsidering, zhou2021estimating1,zhou2021estimating2} is also deeply connected to counterfactual inference while the former focuses on the differences between expected outcomes over the population before and after intervention. We present more ITE works in the supplement.

\vspace{-2mm}
\paragraph{Quantile Regression}  
Traditional regression estimation focuses on estimating the conditional mean of $Y$ given $X$, typically represented by the function $f(X)$. On the other hand, quantile regression \citep{koenker2001quantile} is concerned with estimating conditional quantiles, specifically the $\tau$-th quantile $\mu_{\tau}$, which is the minimum value $\mu$ such that $P(Y \leq \mu|X)=\tau$, where $\tau$ is a predefined value. Some quantile regression settings encounter the quantile crossing problem \citep{takeuchi2009nonparametric}, and \citet{tagasovska2019single} proposes a loss function to learn all the conditional quantiles of a given target variable with neural networks to address this issue. Fortunately, as our framework only requires learning a single quantile for the observation which contains important information about the noise term, this issue can be avoided. 

\vspace{-1mm}
\section{Quantile-Regression-based Counterfactual Inference: Theoretical Insights}
\label{sec:identifiability}
\vspace{-1.5mm}

Conventional counterfactual inference approaches typically rely on estimating both structural causal models and noise values. However, the simultaneous estimation of these elements can be challenging. 
In this paper, we establish a novel connection between counterfactual inference and quantile regression, inspired from the reinforcement learning literature \citet{lu2020sample}.
This connection allows us to bypass the need for estimating both structural causal models and noise values. 
Specifically, the counterfactual outcome $Y_{X = x'} | X=x, Y=y, Z=z$ is equal to the $\tau$-th quantile of the conditional distribution $P(Y|X=x', Z=z)$, with $Y=y$ representing the $\tau$-th quantile of $P(Y|X=x, Z=z)$. Exploiting this connection, the counterfactual outcome can be directly estimated through quantile regression alone. This fundamental relationship is formalized in the theorem below.

\begin{Theorem}
 \label{Theorem}
   Suppose a random variable $Y$ satisfies the following structural causal model:
  \begin{equation*}
  \vspace{-1mm}
      Y = f(X, Z, E)
      \vspace{-.5mm}
  \end{equation*}
   where $X$ and $Z$ cause $Y$, with $Z$ being a cause to $X$. $E$ is the noise term, indicating some unmeasured factors that influence $Y$, with $E \independent X,Z$. 
   We assume that 
   $f$ (which is unknown) is smooth and strictly monotonic in $g(E)$ for fixed values of $X, Z$. Suppose we have observed $\langle X = x, Y = y, Z = z \rangle$. Then for the counterfactual inquiry, what would be the outcome of $Y$ if $X$ had been set to $x'$, given the observed evidence $(X=x, Y=y, Z=z)$, the counterfactual outcome $Y_{X = x'} | X=x, Y=y, Z=z$ is equal to the $\tau$-th quantile of the conditional distribution $P(Y|X=x', Z=z)$, where $Y=y$ represents the $\tau$-th quantile of $P(Y|X=x, Z=z)$.
\end{Theorem}

The theorem establishes a connection between counterfactual outcomes and quantiles, providing the groundwork for identifying counterfactual outcomes through quantile regression from purely factual observations under mild assumptions. In particular, it relies on the assumption that the function $f$ is strictly monotonic w.r.t. $g(E)$, where $g(\cdot)$ is any arbitrary function of the latent factor $E$. Note that, equivalently, $Y = f(X, Z, E)$ can be represented as $f_1(f_2(X,Z), g(E))$.  Furthermore, this theorem additionally brings the following two important theoretical and practical advantages compared to existing approaches for counterfactual inference.

\textbf{Counterfactual outcome is identifable even when $f$ is not identifiable}. 
Numerous methods have attempted to identify the true generating function $f$ and infer the noise value by inverting the estimated function. However, the structural causal model may not always be identifiable \citep{zhang2015estimation}. In contrast, even if the structural causal model lacks identifiability, quantile regression can still be employed to identify the counterfactual outcome under our monotonicity assumption. Additionally, note that the theorem remains valid regardless of whether the variables involved are continuous or discrete. 

\textbf{No need to recover the true noise $E$}. An empirical approach for counterfactual inference based on the above theorem will be introduced in Section \ref{Sec:approach}, which avoids the estimation of the value of the noise term. Moreover, introducing monotonicity on $g(E)$, rather than directly on $E$, not only relaxes the constraints on structural causal models but also on the distribution of the unobserved term. In contrast, recent counterfactual methods such as BGM \cite{nasr2023counterfactual2} and DCM \cite{chao2023interventional}, although considering an invertible function of the noise term, often assume a specific distribution for the noise term (e.g., Gaussian) and rely on the estimation of such noise values. This assumption makes them inflexible when encountering different forms of $g$. 

\begin{figure}
\centering
\begin{subfigure}{0.23\textwidth} 
    \includegraphics[width=3.3cm, height=2.4cm]{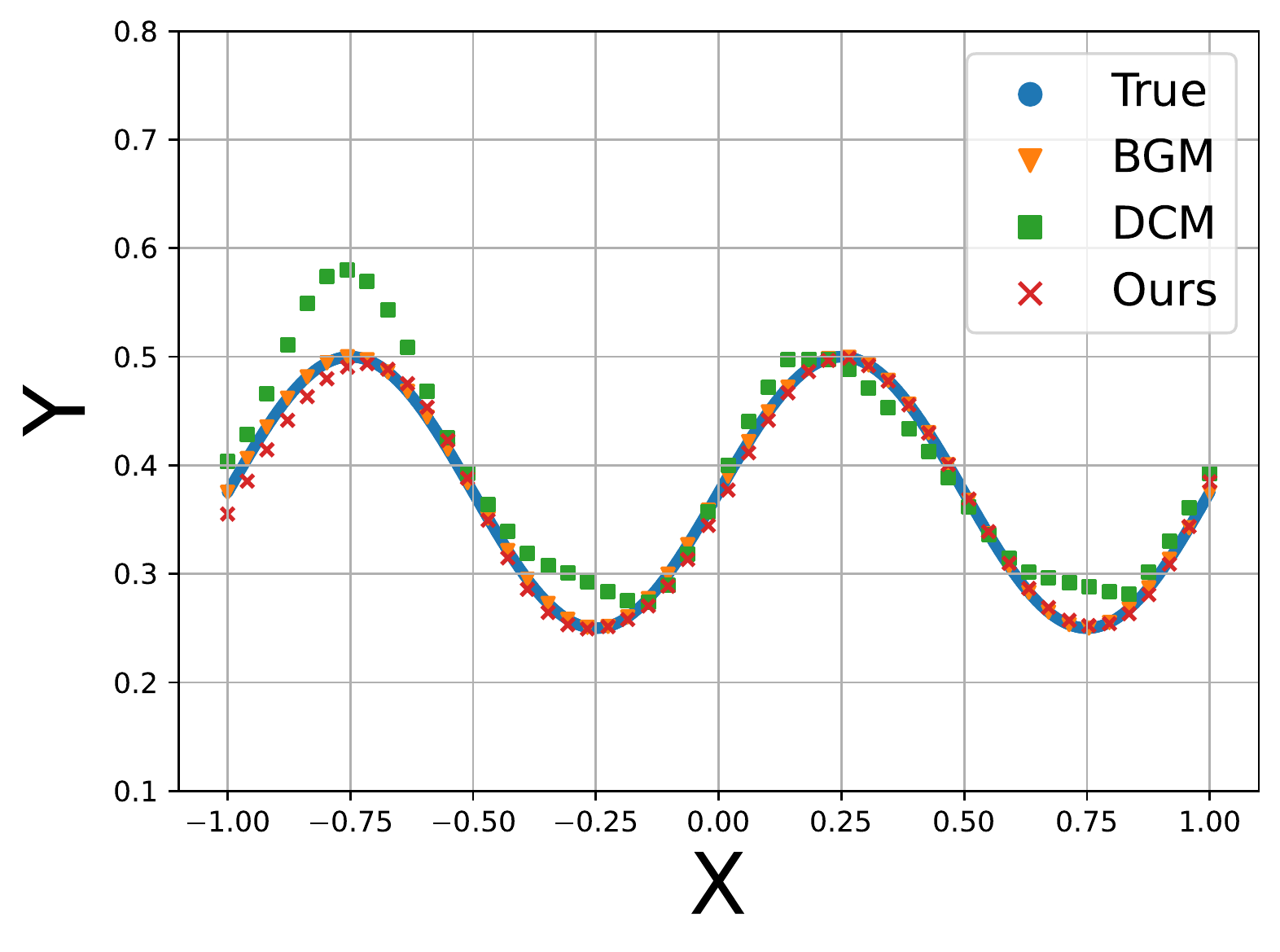}
    \setlength{\abovecaptionskip}{0pt}
    \caption{\small $g(E)=E$.}
    \label{fig:first}
\end{subfigure}
\begin{subfigure}{0.24\textwidth}
    \includegraphics[width=3.5cm, height=2.4cm]{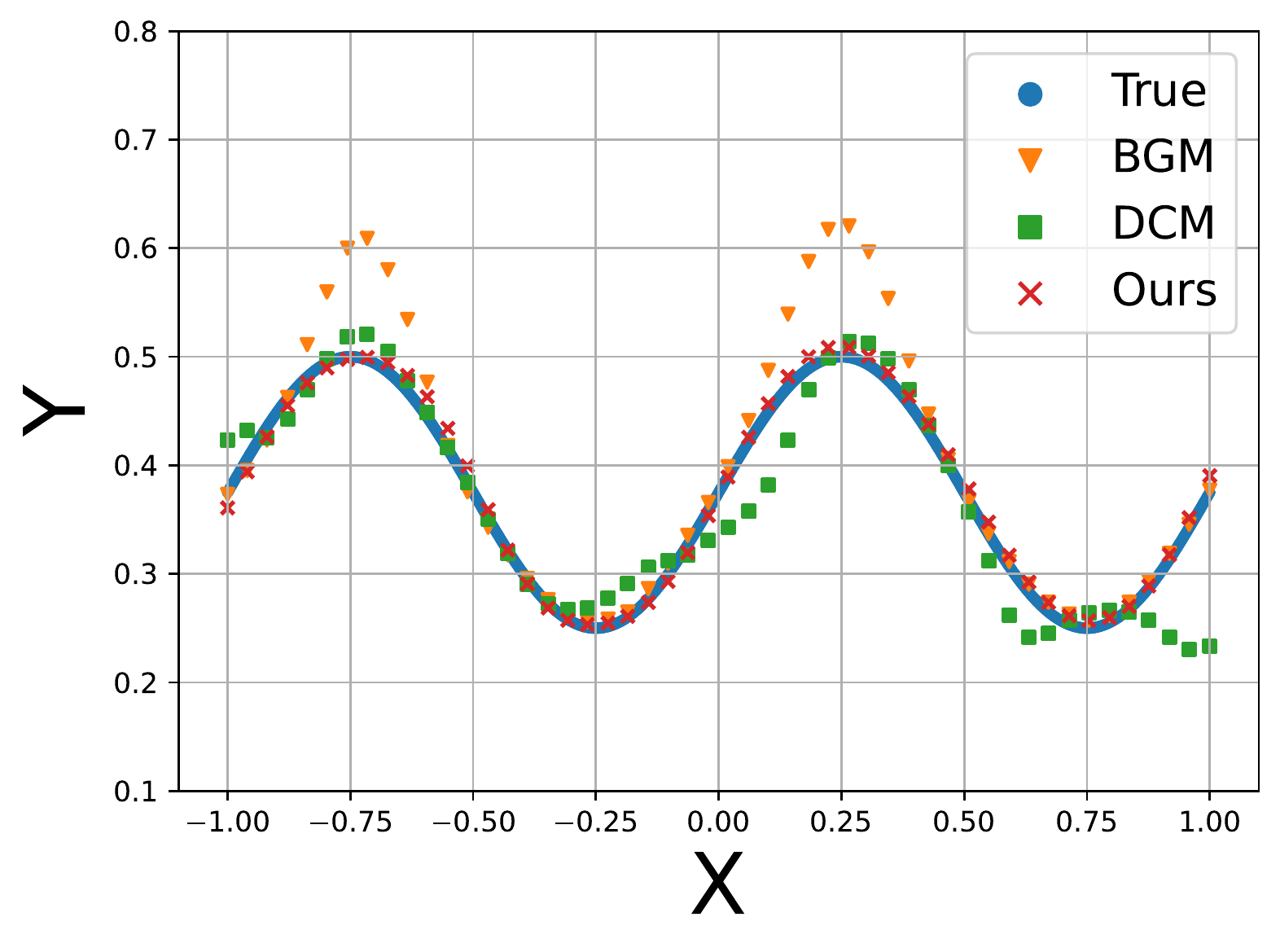}
    \setlength{\abovecaptionskip}{0pt}
    \caption{\small $g(E)=min(max(E, -0.5),1)$}
    \label{fig:second}
\end{subfigure}
\vspace{-0.3cm}
\caption{\small Comparisons of the counterfactual predictions when traversing $X$ under two different forms of $g(E)$. Our method, based on Theorem \ref{Theorem}, demonstrates resilience to various forms of $g(E)$. }
\vspace{-0.4cm}
\label{fig:ge_illus}
\end{figure}

For instance, consider the following two scenarios: in the first scenario, let $Y=\frac{1}{4}(Z+\sin(2\pi X)g(E)+2g(E))$, $E$ satisfies a Gaussian distribution, and $g(E)=E$, while in the second scenario, $g(E)=\min(\max(E, -0.5),1)$. For a particular sample $\langle X=0.5, Z=0.5 \rangle$ (the unknown noise $E=0.5$), we have $Y=1.5$ for both scenarios. Now we estimate the counterfactual outcome when $X$ had been set to different values, with results illustrated in Figure \ref{fig:ge_illus}. We can see our approach, which relies on Theorem \ref{Theorem}, remains unaffected by the form of $g(E)$, since in both scenarios, it relies on $P(Y\leq 1.5|X=0.5, Z=0.5)$. However, BGM and DCM exhibit vastly different performances in the two scenarios because they assume Gaussian noise terms, while it is impossible to recover a noise term that is both Gaussian and satisfies invertibility when $g(E)=\min(\max(E, -0.5),1)$. 

Theorem \ref{Theorem} relies on the monotonicity condition on $g(E)$ and covers a wide range of cases. 
Below, we present a compilation of commonly encountered special cases where this condition remains valid.
\begin{itemize}[leftmargin=10pt,topsep=-3pt,itemsep=-3pt]
    \item Linear causal models: $Y = aX + bZ + g(E)$.
    \vspace{-0.06cm}
    \item Nonlinear causal models with additive noise: $Y=f(X,Z)+g(E)$.
     \vspace{-0.06cm}
    \item Nonlinear causal models with multiplicative noise: $Y=f(X,Z) \cdot g(E)$.
     \vspace{-0.06cm}
    \item Post-nonlinear causal models: $Y=h(f(X,Z)+ g(E))$.
     \vspace{-0.06cm}
    \item Heteroscedastic noise models: $Y=f(X,Z) + h(X,Z) \cdot g(E)$.
\end{itemize}

\vspace{-2mm}
\section{Quantile-Regression-based Counterfactual Inference: Practical Approaches}\label{Sec:approach}
\vspace{-1mm}

Building on the theoretical results presented in Section \ref{sec:identifiability}, we have established that counterfactual inference can be reframed as an extended quantile regression problem. This reformulation eliminates the requirement to estimate the structural causal model and noise values for addressing counterfactual inquiries. Accordingly, in this section, we introduce a practical approach for counterfactual inference with quantile regression and establish an upper bound on the generalization error. 

In particular, in Section \ref{Sec:bilevel}, we formulate counterfactual inference under a bi-level optimization scheme. The upper level is dedicated to estimating the targeted quantile level $\tau$, while the lower level endeavors to estimate the quantile regression function at the specific quantile level. To solve this bi-level optimization problem, Section \ref{Sec:NN} introduces a practical estimation approach that employs neural networks capable of accommodating general causal models and data distributions. Additionally, for more efficient inference of multiple counterfactual samples, we adopt a compact representation to encapsulate the variation in quantile regression functions for different samples, eliminating the need to retrain the bilevel optimization for each new sample. Furthermore, Section \ref{Sec:generalization} provides an in-depth analysis of the generalization ability of the proposed approach and establishes an upper bound on the generalization error. 


\vspace{-1mm}
\subsection{Counterfactual Inference under a Bi-Level Optimization Scheme}\label{Sec:bilevel}
\vspace{-.5mm}


Suppose there are $N$ samples $\{x_i, y_i, z_i\}_{i=1}^N$ which are realizations of random variables $X$, $Y$, and  $Z$. We are interested in finding the counterfactual outcome $y^{\prime}$ which is the realization of $Y_{X=x^\prime}$ for a particular sample of interest $\langle x, y, z \rangle$. An illustration of our method is provided in Fig. \ref{fig:illus}: we first estimate the quantile $\tau=P(Y\leq y|x, z)$ and its corresponding quantile function $\mu_{\tau}(X, z)=\min_{\mu}[P(Y\leq \mu|X,z)=\tau]$. Then we can infer the counterfactual outcome with $\mu_{\tau}(x^\prime, z)$.

However, estimating $\tau$ and $\mu_{\tau}$ remains challenging. A straightforward way is to estimate $P(Y\leq y|X=x, Z=z)$ as $\tau$ first, e.g., with Monte Carlo, and then perform standard quantile regression to obtain the corresponding quantile function. However, $P(Y\leq y|x, z)$ can be difficult to estimate with finite training samples. For example, there may be few or even only one training sample (itself) that have $X=x, Z=z$, leading to inaccurate estimation of $\tau$, which has been demonstrated in the experiments in Section \ref{subsec:exp_quantile_learning}.

To address this problem, we couple the estimations of $\tau$ and the quantile function $\mu_{\tau}$, and
formulate the counterfactual inference problem as a bi-level optimization problem: 
\vspace{-.5mm}
\begin{flalign} 
\hat{\tau} &= \arg \min_\tau {L}(\hat{\mu}_{\tau}(x, z),y),  \textrm{with} \\ \nonumber
\hat{\mu}_{\hat{\tau}} &= \arg \min_{\mu} R_{\hat{\tau}}[\mu],
\label{eq:bilevel}
\end{flalign}

\vspace{-4mm}
where the upper level is to estimate the quantile $\tau$ and $L(.,.)$ can be a regular regression loss, such as $L_1$ and $L_2$ loss.
As for the lower level problem, it is a standard quantile regression problem for the particular quantile level, with the quantile regression function:
\vspace{-3mm}
\begin{flalign}
&R_{\tau}[\mu] = \frac{1}{N} \sum\nolimits_{i=1}^N l_{\tau}(y_i - \mu(x_i, z_i))\\
 & \text{and } l_{\tau}(\xi)=\begin{cases}
  \tau \xi, &\text{ if } \xi \geq 0 \\
  (\tau-1) \xi, & \text{ if } \xi < 0. 
\end{cases}  
\label{eq:pin_ball}
\vspace{-2mm}
\end{flalign}

\begin{figure*}
    \centering
    \includegraphics[scale=0.62]{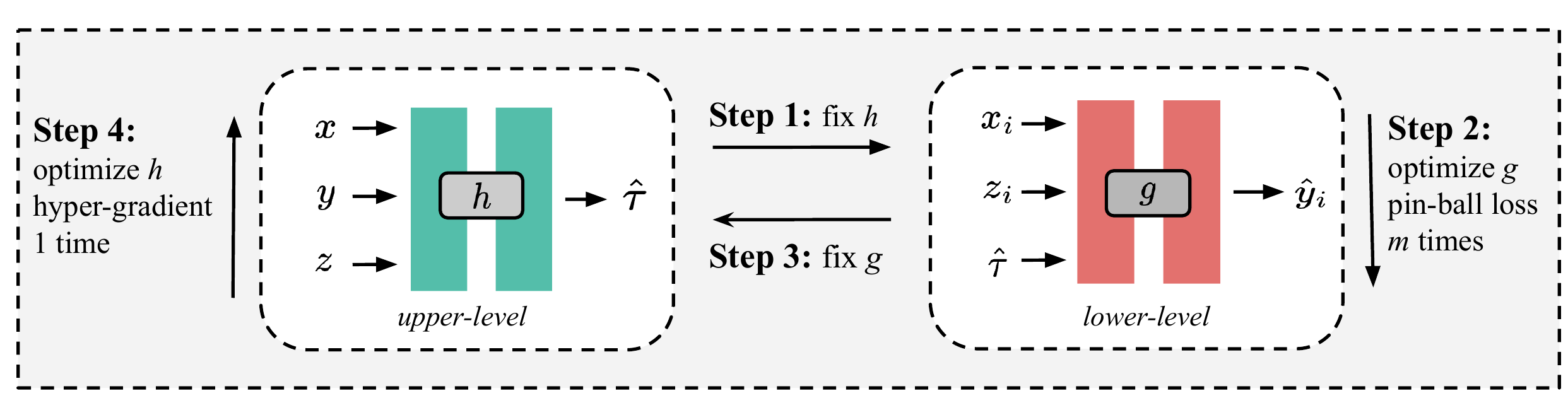}
    \vspace{-2mm}
    \caption{\small One training step of our proposed bi-level implementation.}
    \vspace{-4mm}
    \label{fig:diagram}
\end{figure*}

\vspace{-2mm}
We use the pin-ball loss $l_{\tau}$ as the objective as it has been shown that the minimizer of the empirical pin-ball loss converges to the true quantile $\mu_{\tau}$ under some mild assumptions \citep{takeuchi2006nonparametric}. 

After the bi-level optimization, we obtain an accurate quantile regressor $\hat{\mu}_{\tau}$.
Since the factual outcome $y$ is the $\tau$-th quantile of the $P(Y|x, z)$, $\tau$ is the minimizer of the objective $L(\hat{\mu}_{\tau}^*(x, z),y)$. \!In other words, we have $\hat{\tau}\!\!=\!\!\tau\!\!=\!\!P(Y\leq y|x,z), \hat{\mu}_{\hat{\tau}}\!=\!\mu_{\tau}$.
Through the lens of the above bi-level optimization formulation, we avoid the direct estimation of $P(Y\leq y|x, z)$ with finite samples. Next, we show how to efficiently solve this bi-level optimization problem in practice.


\vspace{-2mm}
\subsection{An Efficient Neural-Network-Based Implementation} \label{Sec:NN}
\vspace{-1.5mm}
Although we have identifiability guarantees for the counterfactual outcome, it still remains unclear how should we implement the framework to solve the bi-level optimization problem.  In this context, we present a scalable and efficient neural network-based implementation, for both lower-level and upper-level optimization. We employ a neural network $h$ to learn the quantile of the sample and network $g$ to learn the quantile function. A training pipeline is given in Fig. \ref{fig:diagram}. 



\vspace{-2mm}
\paragraph{Lower-level optimization.} 

Since each sample of interest $x$ corresponds to a different quantile $\tau$ and quantile function $\mu_{\tau}$, the computational cost can be huge if we learn the quantile regression function $\hat{\mu}_{\hat{\tau}}$ for every interested sample separately. Hence, to achieve efficient quantile estimation of counterfactual inference for multiple samples, we propose to learn a conditional neural network $g$. Specifically, we use $\hat{\tau}$ to capture the difference in different samples and concatenate $\hat{\tau}$ with each training sample $\{x_i, z_i\}_{i=1}^N$ as input to the network $g$ as shown in the right part of Fig. \ref{fig:diagram}. It eliminates the need to retrain the bilevel optimization for each sample of interest: 
\begin{equation*}
\vspace{-2mm}
    \hat{\mu}_{\hat{\tau}}(x_i,z_i) \Longrightarrow g(x_i,z_i, \hat{\tau}).
\end{equation*}
Then for each $\hat{\tau}$, we minimize the pin-ball loss $\frac{1}{N}\sum_{i=1}^N l_{\hat{\tau}}(y_i-g(x_i, z_i, \hat{\tau}))$ following Eq. \ref{eq:bilevel}.
 Accordingly, after the optimization procedure, we have the $\hat{\tau}$-th quantile regression output as $g(x, z, \hat{\tau})$ for every sample of interest $\langle x, z, y \rangle$ with a shared neural network $g$.


\vspace{-2mm}
\paragraph{Upper-level optimization.} 
To recover the quantile $P(Y\leq y|X=x, Z=z)$ from the factual observations $\langle x, y, z\rangle$, we propose a data-dependent model to learn the quantiles for each sample of interest automatically.
Specifically, we employ a neural network $h$ to infer $\tau$ from the observational data, i.e., $\hat{\tau}=h(x, z, y)$, and use $\hat{\tau}$ as the input of the network $f$ in the lower-level problem to perform quantile regression. 
An important advantage of the data-dependent implementation is that it allows inferring $\tau$ and counterfactual outcomes for an unseen data sample $\langle x, z, y\rangle$. For example, we may infer the $\tau$ for samples in the validation split even though they have not been used in the training.
The optimization of the upper-level problem is more challenging 
as we also need to consider the lower-level constraint besides the upper-level regression loss $L(.,.)$. Thanks to the multi-level optimization library--Betty \citep{choe2023betty},  we are able to solve this bi-level optimization problem without considering the complex interactions between the lower and upper-level problems. It 
uses hyper-gradient \citep{choe2023making} to update the network $h$ automatically.

\textbf{Summary of Training Pipeline}. We present the training pipeline of one step in Fig. \ref{fig:diagram}. In each training step, we have four sub-steps: 1) fix the estimated quantiles (i.e., network $h$); 2) train the neural network $g$ with the pin-ball loss for each $\hat{\tau}$ by stochastic gradient descent (SGD) $m$ times ($m=30$ in our experiments); 3) send the optimized network $g$ to the upper-level problem. 4) the bi-level optimization library updates the network $h$ one time by hyper-gradient to minimize the regression loss $L(.,.)$. Then we continue the training steps until the models converge. 



\vspace{-0.2cm}
\subsection{Generalization Bound of Empirical Estimator}\label{Sec:generalization}
\vspace{-0.1cm}
Ideally, the counterfactual outcome $Y_{x^\prime}|X=x,Z=z, Y=y$ is $\mu_{\tau}(x', z)$.
However, we are usually given a limited number of training samples and the pair $(\tau, \mu_{\tau})$ is approximated by using the bi-level optimization solution $(\hat{\tau},\hat{\mu}_{\hat{\tau}})$. 
For a sample of interest $\langle x,y,z\rangle$, we have the estimated quantile $\hat{\tau}=h(x,y,z)$ and counterfactual predictions $y^\prime=\hat{\mu}_{\hat{\tau}}(x^\prime, z)=g(x^\prime,z, \hat{\tau})$ for any new value $x^\prime$. An essential problem is that we are unsure about the generalization ability of the regressor $\hat{\mu}_{\hat{\tau}}$ on the counterfactual input pair $\langle x^\prime, z \rangle$ as the pair has not been seen by $\hat{\mu}_{\hat{\tau}}$ during training. 

\begin{figure*}
    \centering
    \setlength{\tabcolsep}{1pt}
   \begin{tabular}{cccccc}
       \includegraphics[width=3.2cm, height=2.5cm]{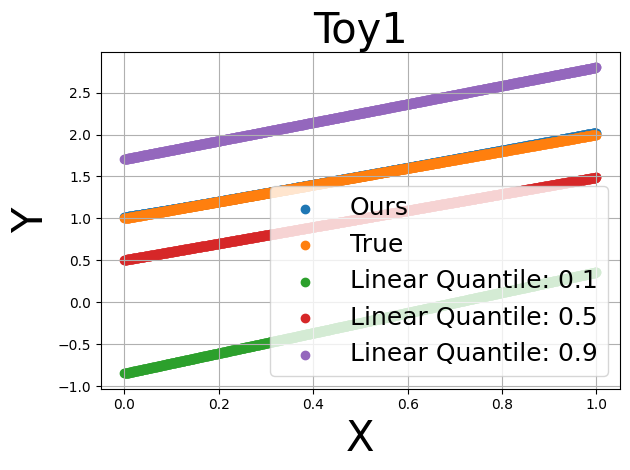} &\includegraphics[width=3.2cm, height=2.5cm]{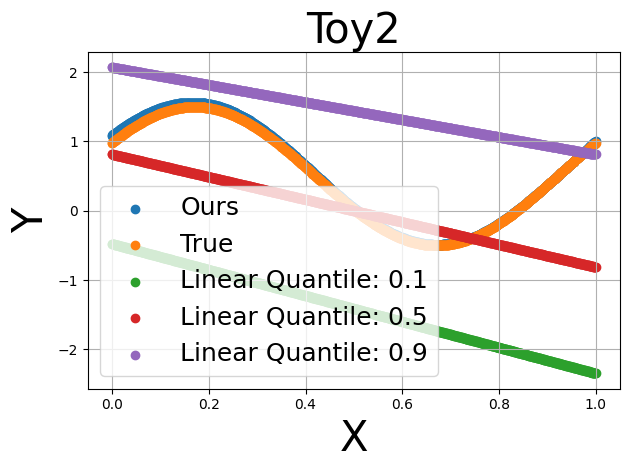}
       &\includegraphics[width=3.2cm, height=2.5cm]{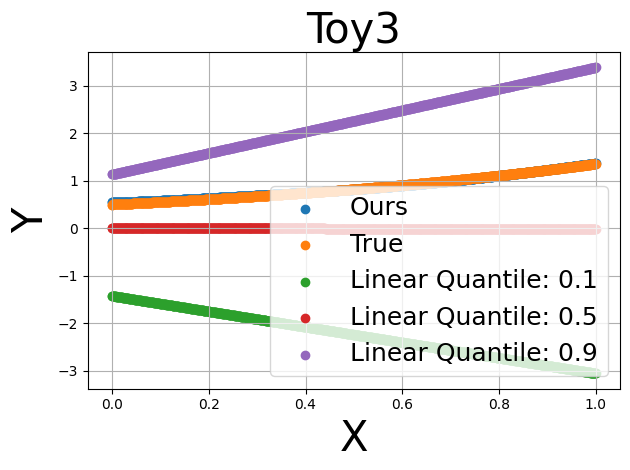}
       &\includegraphics[width=3.2cm, height=2.5cm]{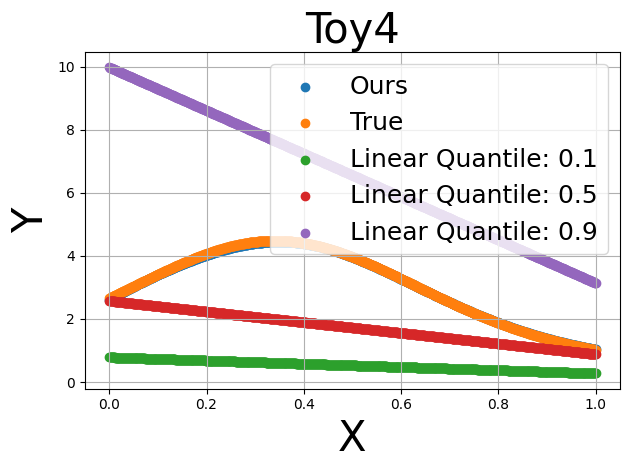}
       &\includegraphics[width=3.2cm, height=2.5cm]{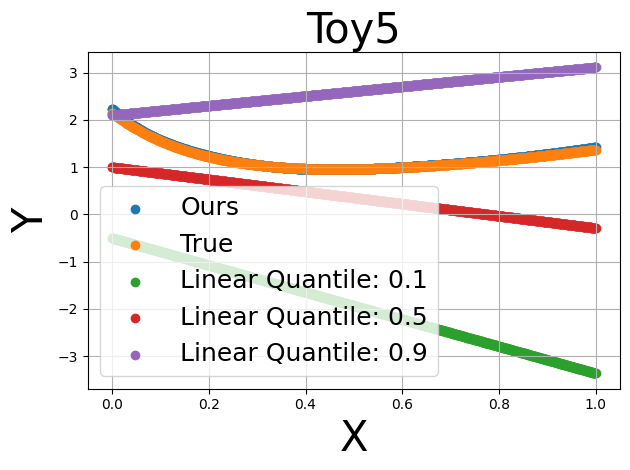}
   \end{tabular}
    \vspace{-0.5cm}
    \caption{\small Toy examples for counterfactual estimations. For the interested sample, we traverse the value of $X$ with $x^\prime$ and compare against with the true counterfactual $Y_{x=x^\prime}$. Our method is able to recover the true trajectory from factual observations under the five causal models. 
    }
     \vspace{-0.3cm}
    \label{fig:toy_counterfactual}
\end{figure*}

\begin{figure}
    \centering
    \includegraphics[scale=0.35]{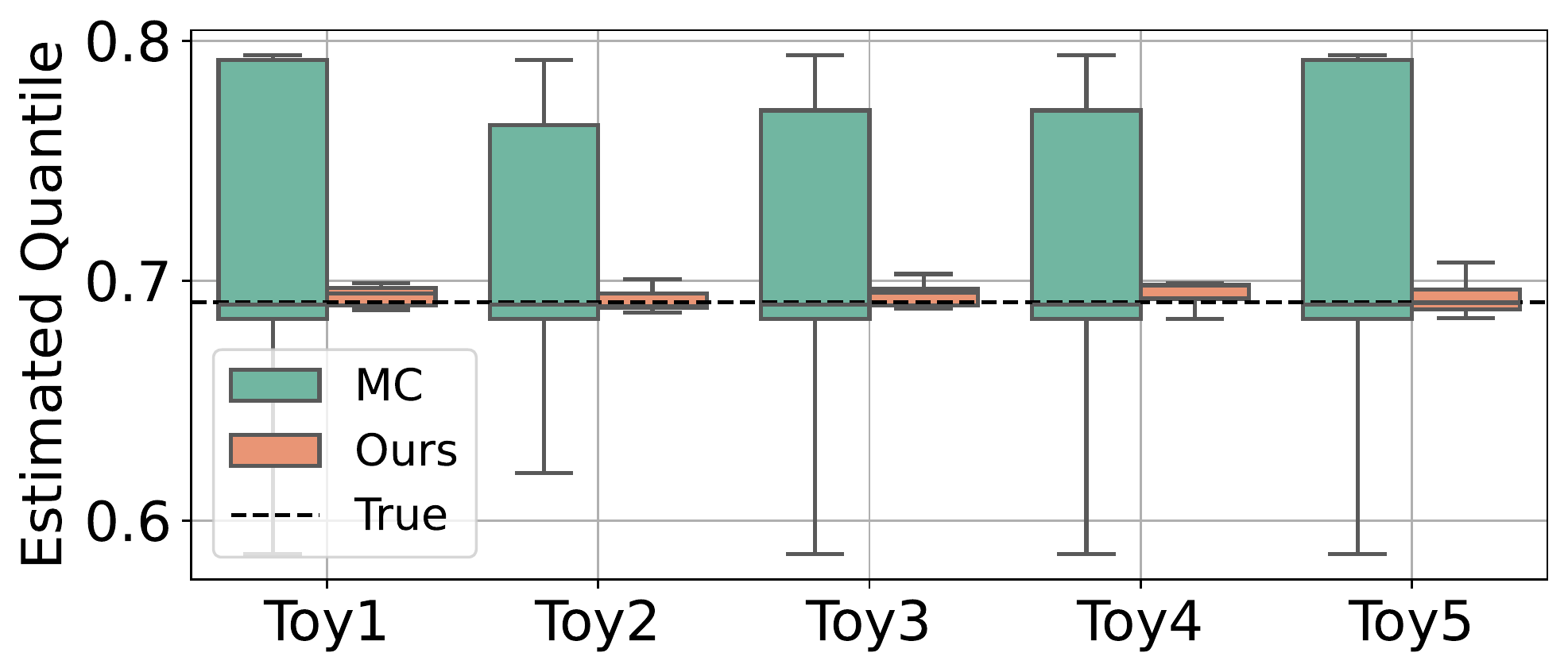}
    \caption{\small Box plots of the learned quantiles. The ground truth quantile is $\Phi(0.5)=0.691$. Compared to the Monte Carlo (MC) method, 
    our proposed bilevel optimization method learns the quantile stably while the MC method has a significantly larger variance, which makes it unsuitable for counterfactual inference. }
    \label{fig:box_quantile}
\end{figure}
More formally, we are interested in analyzing the upper bound of the generalization error $\mathbb{E}_{x,z}[l_{\hat\tau}(\mu_{\hat{\tau}}(x, z)- \hat{\mu}_{\hat{\tau}}(x, z))]$. Below, we employ the Rademacher complexity to upper bound the generalization error.
\begin{Definition}[Rademacher complexity \citep{bartlett2002rademacher}]
Let $F$ be a hypothesis class mapping from $\mathcal{X}$ to $[0,b]$. Let $\{x_i, z_i\}_{i=1}^N$ be i.i.d. examples. Let $\{\sigma_i\}_{i=1}^N$ be independent Rademacher variables taking values in $\{-1,+1\}$ uniformly. The Rademacher complexity is defined as
\vspace{-1mm}
\begin{flalign} \nonumber
\mathfrak{R}(F)=\mathbb{E}_{x,z,\sigma}\left[\sup_{\mu \in F}\frac{1}{N}\sum_{i=1}^N\sigma_i \mu(x_i, z_i)\right].
\end{flalign}
\end{Definition}

\vspace{-1.5mm}
Our main theoretical result is as follows.
\begin{Theorem}\label{theorem:bound}
Let $(\hat\tau,\hat{\mu})\in(\mathfrak{T}, F)$ by the optimization solution, where $\mathfrak{T}$ is the parameter space. Let the loss function $l_\tau$ be upper bounded by $b$. Then, for any $\delta>0$, with probability at least $1-\delta$, we have
\vspace{-2mm}
\begin{flalign} \nonumber
\mathbb{E}_{x,z}[l_{\hat\tau}(\mu_{\hat{\tau}}(x, z) - \hat{\mu}_{\hat{\tau}}(x, z))]&\leq \frac{1}{N}\sum_{i=1}^{N}l_{\hat{\tau}}(\mu_{\hat{\tau}}(x_i, z_i) - \hat{\mu}_{\hat{\tau}}(x_i, z_i))\\ \nonumber
&+ 4\mathfrak{R}(F)+\frac{4b}{\sqrt{N}}+b\sqrt{\frac{\log(1/\delta)}{2N}}.
\end{flalign}
\end{Theorem}
\vspace{-1.5mm}
The Rademacher complexity has been widely used to derive generalization error bounds in the statistical machine learning community \citep{mohri2018foundations}. Its upper bound has also been widely studied. If $F$ is an RKHS and the hypotheses are upper bounded, without any strong assumptions, $\mathfrak{R}(F)\leq O(\sqrt{1/N})$ \citep{bartlett2002rademacher}.

The upper bound of $\mathbb{E}_{x,z}[l_{\hat\tau}(\mu_{\hat{\tau}}(x, z) - \hat{\mu}_{\hat{\tau}}(x, z))]$ heavily relies on the (i) empirical value $\frac{1}{N}\sum_{i=1}^{N}l_{\hat{\tau}}(\mu_{\hat{\tau}}(x_i, z_i) - \hat{\mu}_{\hat{\tau}}(x_i, z_i))$, which can be minimized by utilizing the factual training samples and (ii) the number of training samples $N$. Given a fixed number of training samples, our theoretical results imply that the generalization error is bounded, i.e., the counterfactual predictions will be very close to the ground truth counterfactual outcome. We also empirically show that our method can achieve much better performance given same number of samples compared with other strong baselines in Section \ref{subsec:exp_bound}. In summary, the performance of our proposed quantile regression method is guaranteed with finite samples.


\vspace{-2mm}
\section{Experimental Results}
\vspace{-2mm}
In this section, we begin by introducing the experimental setup including the datasets, evaluation metrics, baseline methods, and implementation details. Then we analyze the learned quantiles under different models and
conduct a comprehensive comparison with state-of-the-art approaches across diverse datasets. Additionally, we also study the sample efficiency of our method, the essential monotonicity assumption, and potential latent confounding cases.

\vspace{-1.5mm}
\subsection{Experiment Setup}
\vspace{-1.5mm}
\textbf{Datasets.} Due to the lack of observations in counterfactual scenarios in the real world, we create the following datasets, which satisfy the monotonicity assumption in our theorem.\ref{Theorem}, to evaluate the performance of our method. In \textit{Cont-Dose} and \textit{Dis-Dose}, the covariate $Z$ is the age of patients,  $X$ represents the (continuous and binary, respectively) dose of the medical treatment, and $Y$ denotes the outcome. Both datasets contain 800 training and 200 testing samples. To further evaluate our method on high-dimensional data, we create the \textit{Rotation-MNIST} and \textit{Thick-Omniglot} datasets. In Rotation-MNIST, $Z$ is the original MNIST images \cite{lecun2010mnist}, $X$ is the rotation of the digit, and noises are the changes to the RGB values of the rotated images. In Thick-Omniglot, $Z$ is the original Omniglot images \cite{lake2019omniglot}, $X$ changes the thickness of the digits, and noises are the darkness of the transformed images. 
We also adopt the semi-synthetic \textit{IHDP} dataset \citep{Hill2011}.

\textbf{Implementation.} We provide the implementation details in the appendix and code in the supplement.

\begin{table*}[ht]
    \centering
    \begin{tabular}{|c|cc|cc|cc|} 
    \hline 
     \multirow{2}{*}{Method}   & \multicolumn{2}{c|}{IHDP} &  \multicolumn{2}{c|}{Cont-Dose} & \multicolumn{2}{c|}{Dis-Dose}\\ \cline{2-7}
    & {Train} & {Test} & {Train} & {Test} & {Train} & {Test}\\ \hline
         DeepSCM \cite{pawlowski2020deep} & 2.37 $\pm$ 2. & 2.98 $\pm$ 4. & 0.38 $\pm$ .0 & 0.40 $\pm$ .0&0.33 $\pm$ .0 & 0.36 $\pm$ .0\\
         CFQP-T \cite{de2022deep} & 1.81 $\pm$.1  & 1.80 $\pm$ .1&0.19 $\pm$ .0 &0.19 $\pm$ .0&0.22 $\pm$ .0 & 0.22 $\pm$ .0\\
         CFQP-U \cite{de2022deep} & 1.40 $\pm$ .1 & 1.30 $\pm$ .0 &0.19 $\pm$ .0 &0.18 $\pm$ .0& 0.22 $\pm$ .0 & 0.22 $\pm$ .0\\
         BGM \cite{nasr2023counterfactual2} & 4.35 $\pm$ .4 &4.89  $\pm$ .5 &0.31 $\pm$ .0 & 0.39 $\pm$ .1 & 0.27 $\pm$.0  & 0.29 $\pm$ .0 \\ 
         DCM \cite{chao2023interventional} & 2.56 $\pm$ 2. & 2.76 $\pm$ 2. &0.19 $\pm$ .0 &0.16 $\pm$ .0  & 0.28 $\pm$ .0 & 0.29 $\pm$ .0\\\hline
         Ours & \textbf{1.29 $\pm$ .3} & \textbf{1.23 $\pm$ .2}& \textbf{0.06 $\pm$ .0} & \textbf{0.06 $\pm$ .0} & \textbf{0.20 $\pm$ .0} & \textbf{0.20 $\pm$ .0} \\ \hline
    \end{tabular}
    \setlength{\abovecaptionskip}{5pt}
    \caption{\small The RMSE performance for counterfactual inference on tabular datasets. Note that we do not have access to the counterfactual outcomes for training split during training. So, we measure the counterfactual inference performances on both training and testing splits.}
    \label{tab:res_tabular}
\end{table*}

\begin{figure*}[ht]
    \centering
    \vspace{-0.3cm}
    \begin{minipage}[b]{0.5\linewidth}
        \centering
        \begin{tabular}{|c|cc|cc|} 
    \hline 
     \multirow{2}{*}{Method} & \multicolumn{2}{c|}{Rotation-MNIST}  & \multicolumn{2}{c|}{Thick-Omniglot}  \\ \cline{2-5}
    & {Train} & {Test} & {Train} & {Test}  \\ \hline
         DeepSCM   & 6.61 $\pm$ .1& 6.55 $\pm$ .1 &11.31 $\pm$ .1 & 11.46 $\pm$ .1\\
         CFQP-T & 2.82 $\pm$ .1 & 2.80 $\pm$ .0 &4.36 $\pm$ .0 & 4.23 $\pm$ .0\\
         CFQP-U  & 1.79 $\pm$ .0 & 1.78 $\pm$ .0 &  3.28 $\pm$ .1 & 3.30 $\pm$ .1\\
         BGM   & 7.53 $\pm$ .2 & 7.52 $\pm$ .1 & 12.07 $\pm$ .2 & 12.27 $\pm$ .1 \\ 
         DCM  &  5.30 $\pm$ .0 & 5.23 $\pm$ .0 & 8.80 $\pm$ .0 & 9.01 $\pm$ .1\\ \hline
         Ours & \textbf{1.54 $\pm$ .1} & \textbf{1.52 $\pm$ .1} & \textbf{2.60 $\pm$ .0} & \textbf{2.96 $\pm$ .1}\\ \hline
    \end{tabular}
        \captionof{table}{\small The RMSE performance for counterfactual inference on image transformation datasets. }
        \label{tab:res_image}
    \end{minipage}
    \hfill 
\begin{minipage}[b]{0.4\linewidth}
        \centering
        \setlength{\tabcolsep}{1.5pt}
      \begin{tabular}{ccccccc}
      Z & Y &  CFQP & BGM&  DCM & Ours & Truth \\
          \includegraphics[scale=.6]{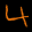} &
\includegraphics[scale=.6]{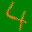} & 
\includegraphics[scale=.6]{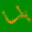} &
\includegraphics[scale=.6]{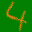} &
\includegraphics[scale=.6]{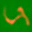} & 
\includegraphics[scale=.6]{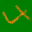} &
\includegraphics[scale=.6]{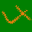}\\
              \includegraphics[scale=0.6]{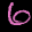} & 
           \includegraphics[scale=0.6]{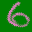} & 
           \includegraphics[scale=0.6]{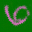}
            & \includegraphics[scale=0.6]{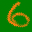}    & \includegraphics[scale=0.6]{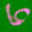}    & \includegraphics[scale=0.6]{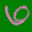}   & \includegraphics[scale=0.6]{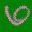}  \\
                \includegraphics[scale=0.6]{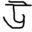} & 
           \includegraphics[scale=0.6]{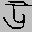} & 
           \includegraphics[scale=0.6]{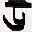}
            & \includegraphics[scale=0.6]{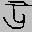}    & \includegraphics[scale=0.6]{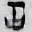}    & \includegraphics[scale=0.6]{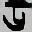}   & \includegraphics[scale=0.6]{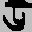}  \\
                \includegraphics[scale=0.6]{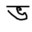} & 
           \includegraphics[scale=0.6]{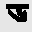} & 
           \includegraphics[scale=0.6]{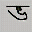}
            & \includegraphics[scale=0.6]{figs/omni/4297_out.png}    & 
            \includegraphics[scale=0.6]{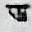}    & \includegraphics[scale=0.6]{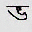}   & \includegraphics[scale=0.6]{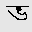}  
      \end{tabular}
      
        \caption{\small Example results on image transformation dataset.}
        \label{tab:image_demo}
    \end{minipage}
    \vspace{-0.3cm}
\end{figure*}

\textbf{Evaluation Metrics and Baseline Methods.} 
We compute the root mean square error (RMSE) between the counterfactual predictions and the ground truth. 
We compare with state-of-the-art baselines: DeepSCM \citep{pawlowski2020deep}, CFQP \citep{de2022deep}, BGM \citep{nasr2023counterfactual2} and DCM \cite{chao2023interventional}. As CFQP provides two backbones, we use CFQP-T to denote the method with a Transformer network and CFQP-U the method with a U-Net backbone. We run the public code of baseline methods with their recommended hyper-parameters.

\vspace{-1mm}
\subsection{Analysis on Quantile Learning}
\label{subsec:exp_quantile_learning}
\vspace{-1mm}
We synthesize the data where the corresponding $\tau$ has closed-form solutions, so we can compare the learned quantiles against the ground truth. Specifically, we consider the following instantiations of the five causal models mentioned in Section \ref{sec:identifiability}, including 1) linear causal models, $Y=X + Z+ E$; 2) nonlinear additive noise model, $Y=\sin(2\pi X+Z)+E$; 3) nonlinear multiplicative noise model, $Y=\exp(X-Z+0.5)\cdot E$; 4) post-nonlinear model, $Y=\exp(\sin(\pi X+Z)+E)$; and 5) heteroscedastic model, $Y=\exp(-5X+Z)+\exp(X+Z-0.5)\cdot  E$. For these five causal models, we have $F^{-1}(\tau)=E \Rightarrow \tau=F(E)$ where $F$ is the CDF of noise $E$. We set $P(X)=P(Z)=\mathcal{U}[0,1]$. As for noise, we consider the isotropic Gaussian distribution $\mathcal{N}(0,1)$.
We sample 100,000 data points for each case. For sample points of interest for counterfactual inference, we use $X=0.5, Z=0.5, E=0.5$ and generate $Y$. Our goal is to learn a quantile for this sample, where the ground-truth quantile $\tau$ is $\Phi(0.5)\approx 0.691$.
\begin{figure*}[ht]
    \centering
   \begin{subfigure}[b]{0.19\textwidth}
         \centering
         \includegraphics[width=3.2cm, height=2.5cm]{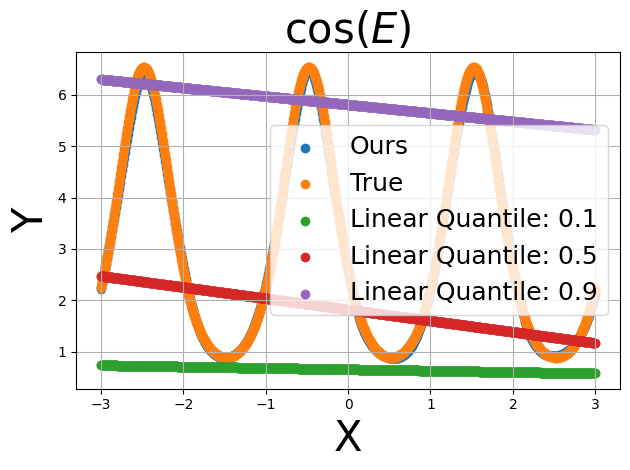}
     \end{subfigure}
      \hfill
     \begin{subfigure}[b]{0.19\textwidth}
         \centering
         \includegraphics[width=3.2cm, height=2.5cm]{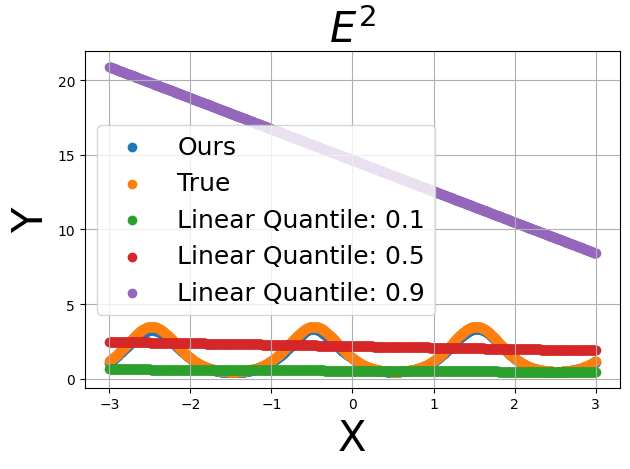}
     \end{subfigure}
     \hfill
     \begin{subfigure}[b]{0.19\textwidth}
         \centering
         \includegraphics[width=3.2cm, height=2.5cm]{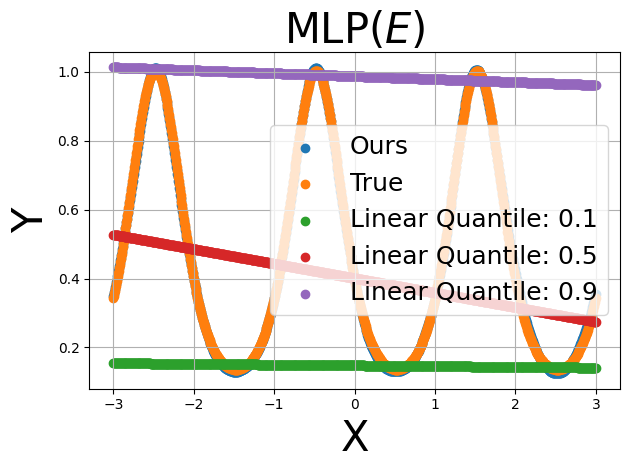}
     \end{subfigure}
     \hfill
     \begin{subfigure}[b]{0.19\textwidth}
         \centering
         \includegraphics[width=3.2cm, height=2.5cm]{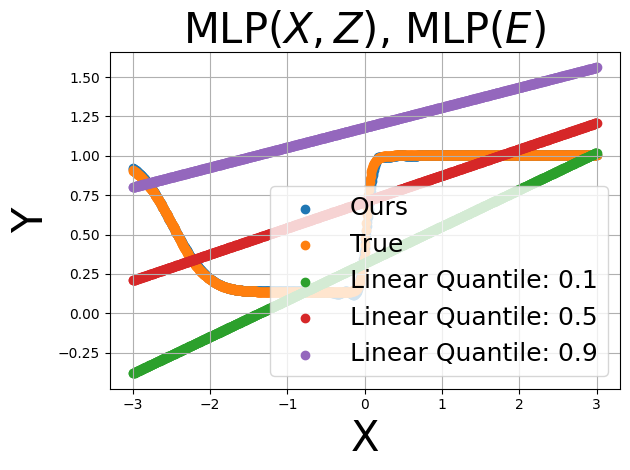}
     \end{subfigure}
     \hfill
     \begin{subfigure}[b]{0.19\textwidth}
         \centering
         \includegraphics[width=3.2cm, height=2.5cm]{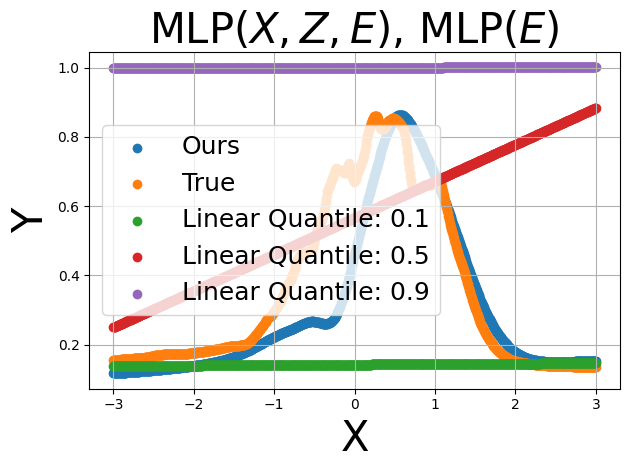}
     \end{subfigure}
     
    \caption{\small Analysis on the monotonicity assumption. (a)$Y=\exp(\cos(\pi X+3Z)+\cos(E))$; (b)$Y=\exp(\cos(\pi X+3Z)+E^2)$; (c)$Y=\exp(\cos(\pi X+3Z)+\text{MLP}(E))$; (d)$Y=\exp(\text{MLP}(X,Z)+\text{MLP}(E))$; (e)$Y=\exp(\text{MLP}(X,Z)+\text{MLP}(E))$.  Although $Y$ is not monotonic w.r.t $E$, the counterfactual outcome is still identifiable since we may have $g(E)=E^2, \cos(E), \text{MLP}(E)$ and $Y$ is monotonic w.r.t $g(E)$. Empirically, we can see that the counterfactual predictions are very close to the ground truth in the first four cases.
    As for the last case (e), it exhibits deviation from the truth since the monotonicity assumption is violated and the counterfactual outcome may not be identifiable.
    }
    
     \vspace{-0.3cm}
    \label{fig:exp_mono}
\end{figure*}
\begin{figure}
    \centering
        \centering
    \includegraphics[width=3.8cm, height=2.8cm]{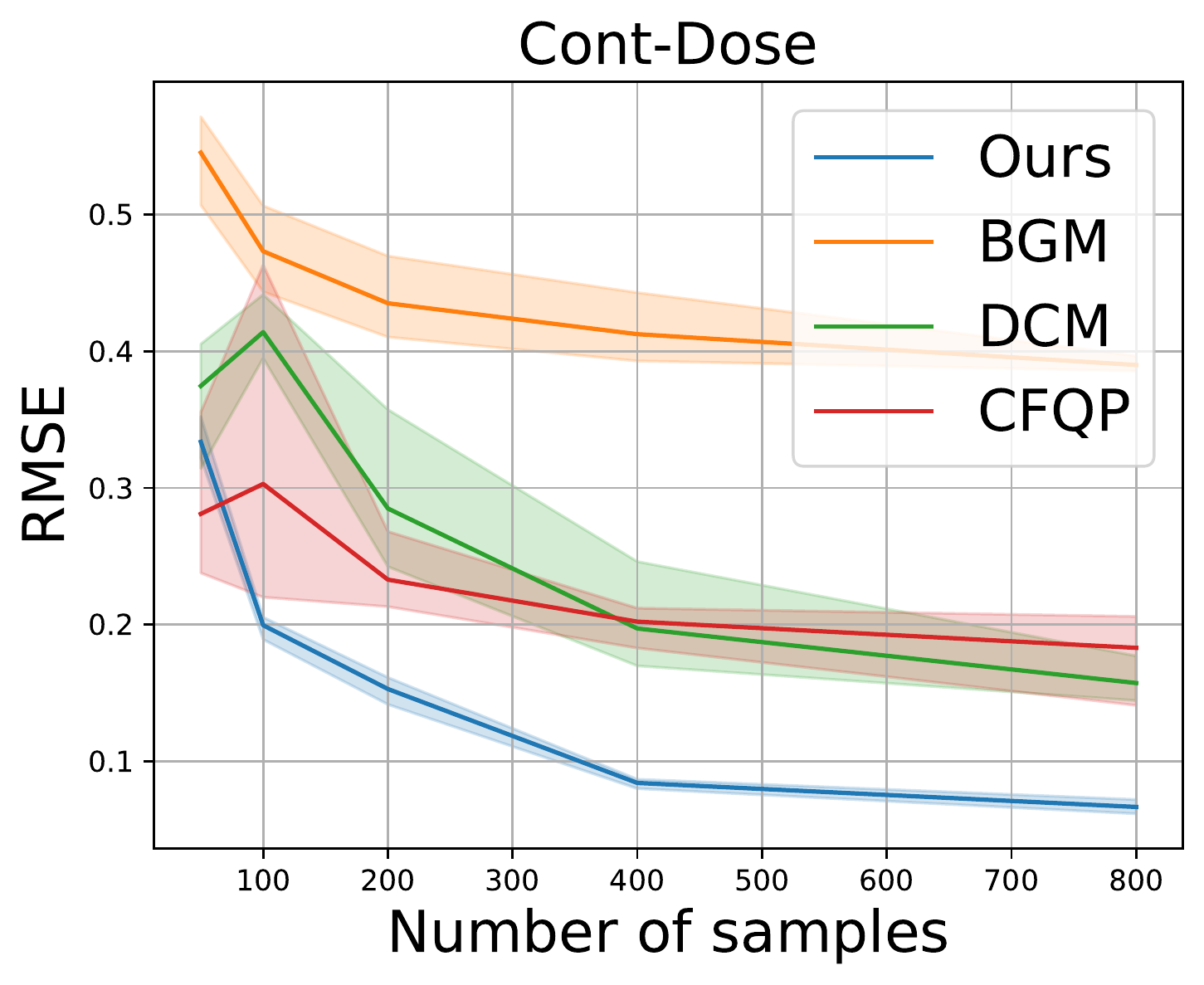}
     ~~~\includegraphics[width=3.8cm, height=2.8cm]{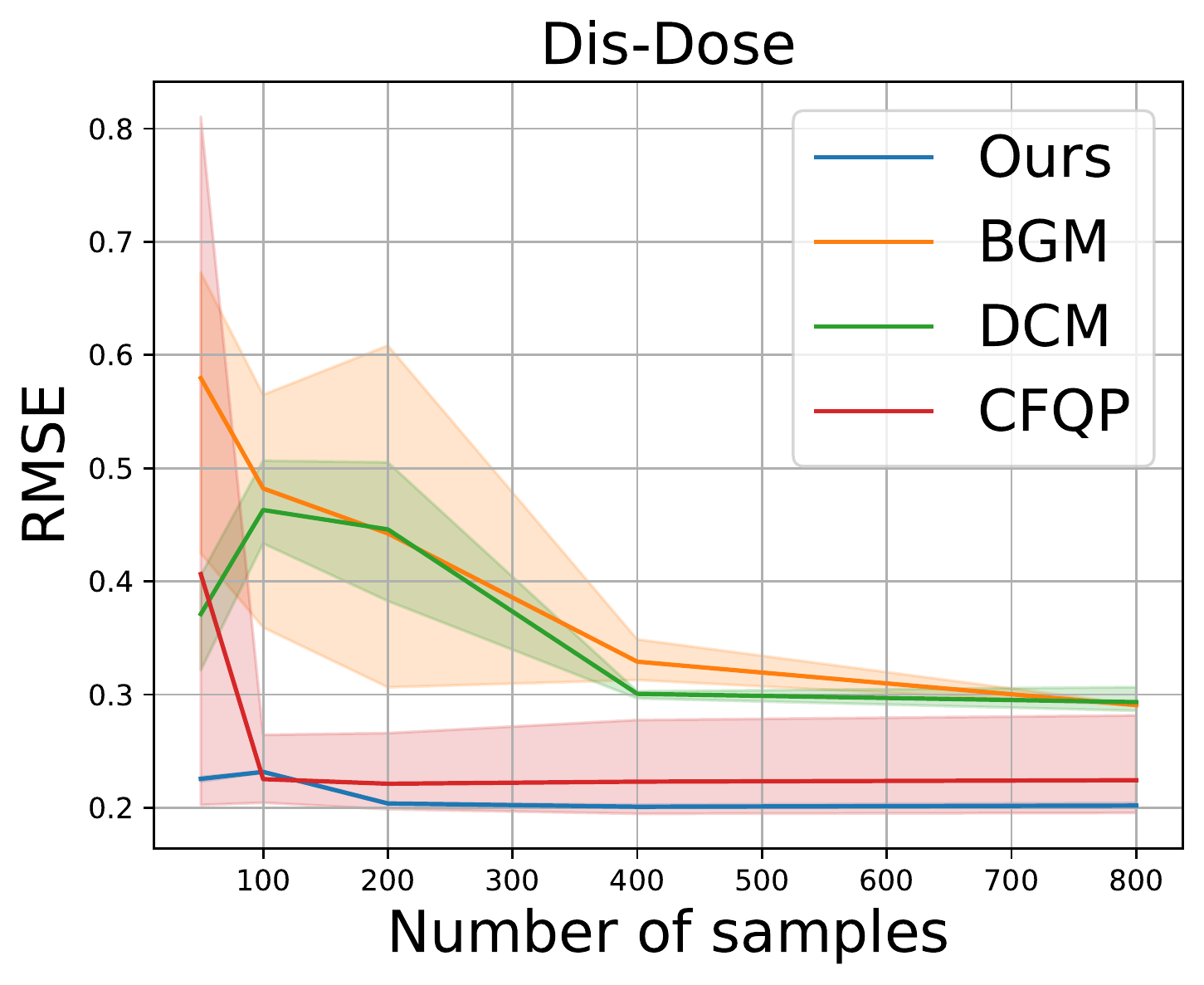}
    \caption{\small The study of sample efficiency.}
    \vspace{-.8cm}
    \label{fig:sample_efficiency}
\end{figure}


First,  as shown in the upper row of Fig. \ref{fig:toy_counterfactual}, we compare with the linear quantile regressor solved by linear programming with specified quantiles at 0.1, 0.5, and 0.9. 
We find that linear quantile regressors are limited by linear parameterization and cannot handle nonlinear generation functions such as the second causal model. Importantly, the baselines do not know the true quantile as they did not leverage the factual observation of interest. By contrast, we observe that our method can achieve accurate counterfactual predictions under all five different causal models, no matter whether they are linear or nonlinear.
Second, one more straightforward way to estimate the quantiles would be Monte Carlo (MC). Specifically, we use $\frac{1}{m }\sum_{i=1}^m \mathbb{I}(y_i\leq y|x_i=x, z_i=z)$ to approximate the quantiles $P(Y\leq y|X=x, Z=z)$, where $m$ is the number of training samples that satisfy $x_i=x,z_i=z$; It is worth noting that MC often failed to find samples with $X=0.5, Z=0.5$ exactly, so we relax it by finding samples with $X$ and $Z$ in range $[0.49, 0.51]$. The problem may be more severe for discrete $X$ and $Z$ since such relaxation is not applicable on discrete variables.
The estimations are visualized in Fig. \ref{fig:box_quantile}. We observe that the variance of learned quantiles in different runs is large with the MC method for all five simple causal models. 
By contrast, our bi-level formulation learns the quantile and quantile function together and estimated quantiles are much more accurate and stable. In particular, our method allows learning the quantile functions for different samples simultaneously by conditioning the estimated quantiles in the lower-level problem, which reduces the training time significantly.

\vspace{-0.15cm}

\subsection{Comparisons with State-of-the-Art Approaches}
 \vspace{-0.08cm}
Table \ref{tab:res_tabular} and Table \ref{tab:res_image} present the results of the counterfactual inference task on tabular and image datasets, respectively. 
Our method achieves the lowest RMSE across all datasets. We observe that the performance of BGM \citep{nasr2023counterfactual2} on image transformation datasets is less than satisfactory. 
BGM employs a conditional spline flow to transform the input to Gaussian noise. However, its expressive power may be limited, especially on high-dimensional image data, compared to our unconstrained neural networks. As a result, it fails to generate accurate counterfactual predictions. 
In Fig. \ref{tab:image_demo}, we present examples of counterfactual image generations. In the third and fourth rows, the strong baseline CFQP \cite{de2022deep} fails to preserve the background color of the outcome image, indicating potential challenges in accurately identifying the noise value within the sample and consequently generating unreliable counterfactual predictions. 
In contrast, our method effectively learns the transformation and preserves the background color in the prediction. This encouraging result further strengthens our identifiability theory.


\subsection{Generalization Bound and Sample Efficiency}
\label{subsec:exp_bound}

We have theoretically shown that the expectation of the generalization error is bounded by the empirical risk and the number of samples.
To verify the theorem, we perform an ablation study on the number of training samples. We work on the Cont-Dose and  Dis-Dose dataset and run experiments with different numbers of samples $N \!\!=\!\! 50, 100, 200, 400, 800$ and report the RMSE averaged over different runs in Fig. \ref{fig:sample_efficiency}. We find that the baselines BGM \citep{nasr2023counterfactual2} and CFQP \cite{pawlowski2020deep} are very sensitive to the number of training samples. In particular, CFQP exhibits a very large variance when the number of samples is smaller than 200. A possible reason is that it assigns samples to different clusters and samples in each cluster become even smaller. In contrast, our method achieves the best performances and the variance is very small, which supports our generalization bounds and demonstrates the sample efficiency of our proposed approach.


\vspace{-2mm}
\subsection{Monotonicity Assumption and Latent Confounder Cases}
\vspace{-0.2cm}
An essential assumption of our method is that there exists a function $g(E)$ such that outcome $Y$ is strictly monotonic w.r.t $g(E)$. Since $g$ can be arbitrary, our assumption covers a very wide spectrum of function classes, as demonstrated by the superior empirical performance of our method. However, it may still be violated in some cases. To investigate this problem, we consider the five causal models as shown in Fig. \ref{fig:exp_mono}. 
In the first four cases, although $Y$ is not monotonic w.r.t $E$, when we set $g(E)=E^2, \cos(E)$, or more complex MLP($E$), $Y$ is monotonic w.r.t $g(E)$. Therefore, according to Theorem \ref{Theorem}, the counterfactual outcome should be identifiable, and the results in the first four cases strongly support this conclusion. As for the last case, the monotonicity assumption is violated and the counterfactual predictions exhibit some deviation from the ground truth, which is expected.  The identifiability of the counterfactual outcome in non-monotonic generation functions remains an important yet challenging task, which we defer to future work.

Similar to \cite{nasr2023counterfactual2,pawlowski2020deep,chao2023interventional}, we also assume no latent confounders. But it would be interesting to testify about the performances under the confounder case. We considered three different confounding cases. 1) $X \leftarrow C\rightarrow Z$; 2) $Z \leftarrow C\rightarrow Y$ and 3) $X\leftarrow C \rightarrow Y$. $C$ is the unknown confounder.
\begin{wraptable}{r}{3.8cm} 
  \begin{tabular}{c|c}
    \hline
    Scenario & RMSE \\ \hline 
    No C & 0.03 $\pm$ .0 \\ \hline
    $X \leftarrow C\rightarrow Z$ & 0.05 $\pm$ .0 \\ \hline
       $Z \leftarrow C\rightarrow Y$ & 0.06 $\pm$ .0   \\ \hline
    $X\leftarrow C \rightarrow Y$   & 0.22 $\pm$ .1   \\ \hline
  \end{tabular}
  \caption{\small Results with the  presence of latent confounder $C$.}
  \vspace{-0.2cm}
  \label{tab:confounder}
\end{wraptable}
The detailed data-generating process is given in the supplementary and the results are shown in Table \ref{tab:confounder}. We observe that our method is quite robust under the first three scenarios, which is reasonable as such $C$ does not affect the causal effect from $X$ to $Y$. 
However, the performance becomes unsatisfactory under the scenario $X\leftarrow C \rightarrow Y$, suggesting that the existence of latent confounders behind $X$ and $Y$ potentially hamper the performances of our method. Extending our approach to cover such latent confounders will be our future work. 

\vspace{-0.15cm}
\section{Conclusion}
Counterfactual inference remains a significant challenge due to the lack of counterfactual outcomes in the real world. In this paper, we advance counterfactual inference from a novel perspective through quantile regression. We first build a connection between counterfactual outcomes and quantile regression and show that the counterfactual outcome corresponds to certain quantities under rather mild conditions. Accordingly, we propose an effective and efficient estimation approach based on neural-network implemented bi-level optimization and show the generalization bound of the empirical estimator. We verify our method on multiple simulated and semi-real-world datasets. The superior performance over state-of-the-art baselines highlights the effectiveness of our method.

\section*{Impact Statement}
This paper presents work whose goal is to advance the field of Machine Learning. There are many potential societal consequences of our work, none which we feel must be specifically highlighted here.


\bibliography{arxiv}

\begin{thebibliography}{41}
\providecommand{\natexlab}[1]{#1}
\providecommand{\url}[1]{\texttt{#1}}
\expandafter\ifx\csname urlstyle\endcsname\relax
  \providecommand{\doi}[1]{doi: #1}\else
  \providecommand{\doi}{doi: \begingroup \urlstyle{rm}\Url}\fi

\bibitem[Bartlett \& Mendelson(2002)Bartlett and Mendelson]{bartlett2002rademacher}
Bartlett, P.~L. and Mendelson, S.
\newblock Rademacher and gaussian complexities: Risk bounds and structural results.
\newblock \emph{Journal of Machine Learning Research}, 3\penalty0 (Nov):\penalty0 463--482, 2002.

\bibitem[Bica et~al.(2020)Bica, Jordon, and van~der Schaar]{bica2020estimating}
Bica, I., Jordon, J., and van~der Schaar, M.
\newblock Estimating the effects of continuous-valued interventions using generative adversarial networks.
\newblock \emph{Advances in Neural Information Processing Systems}, 33:\penalty0 16434--16445, 2020.

\bibitem[Chao et~al.(2023)Chao, Bl{\"o}baum, and Kasiviswanathan]{chao2023interventional}
Chao, P., Bl{\"o}baum, P., and Kasiviswanathan, S.~P.
\newblock Interventional and counterfactual inference with diffusion models.
\newblock \emph{arXiv preprint arXiv:2302.00860}, 2023.

\bibitem[Choe et~al.(2023{\natexlab{a}})Choe, Mehta, Ahn, Neiswanger, Xie, Strubell, and Xing]{choe2023making}
Choe, S.~K., Mehta, S.~V., Ahn, H., Neiswanger, W., Xie, P., Strubell, E., and Xing, E.
\newblock Making scalable meta learning practical.
\newblock \emph{arXiv preprint arXiv:2310.05674}, 2023{\natexlab{a}}.

\bibitem[Choe et~al.(2023{\natexlab{b}})Choe, Neiswanger, Xie, and Xing]{choe2023betty}
Choe, S.~K., Neiswanger, W., Xie, P., and Xing, E.
\newblock Betty: An automatic differentiation library for multilevel optimization.
\newblock In \emph{The Eleventh International Conference on Learning Representations}, 2023{\natexlab{b}}.
\newblock URL \url{https://openreview.net/forum?id=LV_MeMS38Q9}.

\bibitem[De~Brouwer(2022)]{de2022deep}
De~Brouwer, E.
\newblock Deep counterfactual estimation with categorical background variables.
\newblock \emph{Advances in Neural Information Processing Systems}, 35:\penalty0 35213--35225, 2022.

\bibitem[De~Lara et~al.(2021)De~Lara, Gonz{\'a}lez-Sanz, Asher, and Loubes]{de2021transport}
De~Lara, L., Gonz{\'a}lez-Sanz, A., Asher, N., and Loubes, J.-M.
\newblock Transport-based counterfactual models.
\newblock \emph{arXiv preprint arXiv:2108.13025}, 2021.

\bibitem[Du et~al.(2021)Du, Sun, Duivesteijn, Nikolaev, and Pechenizkiy]{du2021adversarial}
Du, X., Sun, L., Duivesteijn, W., Nikolaev, A., and Pechenizkiy, M.
\newblock Adversarial balancing-based representation learning for causal effect inference with observational data.
\newblock \emph{Data Mining and Knowledge Discovery}, 35\penalty0 (4):\penalty0 1713--1738, 2021.

\bibitem[Hill(2011)]{Hill2011}
Hill, J.~L.
\newblock Bayesian nonparametric modeling for causal inference.
\newblock \emph{Journal of Computational and Graphical Statistics}, 20(1), 2011.

\bibitem[Javaloy et~al.(2023)Javaloy, S{\'a}nchez-Mart{\'\i}n, and Valera]{javaloy2023causal}
Javaloy, A., S{\'a}nchez-Mart{\'\i}n, P., and Valera, I.
\newblock Causal normalizing flows: from theory to practice.
\newblock \emph{arXiv preprint arXiv:2306.05415}, 2023.

\bibitem[Johansson et~al.(2016)Johansson, Shalit, and Sontag]{johansson2016learning}
Johansson, F., Shalit, U., and Sontag, D.
\newblock Learning representations for counterfactual inference.
\newblock In \emph{International conference on machine learning}, pp.\  3020--3029. PMLR, 2016.

\bibitem[Khemakhem et~al.(2021)Khemakhem, Monti, Leech, and Hyvarinen]{khemakhem2021causal}
Khemakhem, I., Monti, R., Leech, R., and Hyvarinen, A.
\newblock Causal autoregressive flows.
\newblock In \emph{International conference on artificial intelligence and statistics}, pp.\  3520--3528. PMLR, 2021.

\bibitem[Koenker \& Hallock(2001)Koenker and Hallock]{koenker2001quantile}
Koenker, R. and Hallock, K.~F.
\newblock Quantile regression.
\newblock \emph{Journal of economic perspectives}, 15\penalty0 (4):\penalty0 143--156, 2001.

\bibitem[Lake et~al.(2019)Lake, Salakhutdinov, and Tenenbaum]{lake2019omniglot}
Lake, B.~M., Salakhutdinov, R., and Tenenbaum, J.~B.
\newblock The omniglot challenge: a 3-year progress report.
\newblock \emph{Current Opinion in Behavioral Sciences}, 29:\penalty0 97--104, 2019.

\bibitem[LeCun et~al.(2010)LeCun, Cortes, and Burges]{lecun2010mnist}
LeCun, Y., Cortes, C., and Burges, C.
\newblock Mnist handwritten digit database.
\newblock \emph{ATT Labs [Online]. Available: http://yann.lecun.com/exdb/mnist}, 2, 2010.

\bibitem[Li \& Yao(2022)Li and Yao]{li2022contrastive}
Li, X. and Yao, L.
\newblock Contrastive individual treatment effects estimation.
\newblock In \emph{2022 IEEE International Conference on Data Mining (ICDM)}, pp.\  1053--1058. IEEE, 2022.

\bibitem[Lu et~al.(2020{\natexlab{a}})Lu, Huang, Wang, Hern{\'a}ndez-Lobato, Zhang, and Sch{\"o}lkopf]{lu2020sample}
Lu, C., Huang, B., Wang, K., Hern{\'a}ndez-Lobato, J.~M., Zhang, K., and Sch{\"o}lkopf, B.
\newblock Sample-efficient reinforcement learning via counterfactual-based data augmentation.
\newblock \emph{arXiv preprint arXiv:2012.09092}, 2020{\natexlab{a}}.

\bibitem[Lu et~al.(2020{\natexlab{b}})Lu, Tao, Chen, Li, Guo, and Carin]{lu2020reconsidering}
Lu, D., Tao, C., Chen, J., Li, F., Guo, F., and Carin, L.
\newblock Reconsidering generative objectives for counterfactual reasoning.
\newblock \emph{Advances in Neural Information Processing Systems}, 33:\penalty0 21539--21553, 2020{\natexlab{b}}.

\bibitem[Melnychuk et~al.(2022)Melnychuk, Frauen, and Feuerriegel]{melnychuk2022causal}
Melnychuk, V., Frauen, D., and Feuerriegel, S.
\newblock Causal transformer for estimating counterfactual outcomes.
\newblock In \emph{International Conference on Machine Learning}, pp.\  15293--15329. PMLR, 2022.

\bibitem[Mohri et~al.(2018)Mohri, Rostamizadeh, and Talwalkar]{mohri2018foundations}
Mohri, M., Rostamizadeh, A., and Talwalkar, A.
\newblock \emph{Foundations of machine learning}.
\newblock MIT press, 2018.

\bibitem[Nasr-Esfahany \& Kiciman(2023)Nasr-Esfahany and Kiciman]{nasr2023counterfactual2}
Nasr-Esfahany, A. and Kiciman, E.
\newblock Counterfactual (non-) identifiability of learned structural causal models.
\newblock \emph{arXiv preprint arXiv:2301.09031}, 2023.

\bibitem[Nasr-Esfahany et~al.(2023)Nasr-Esfahany, Alizadeh, and Shah]{nasr2023counterfactual}
Nasr-Esfahany, A., Alizadeh, M., and Shah, D.
\newblock Counterfactual identifiability of bijective causal models.
\newblock \emph{arXiv preprint arXiv:2302.02228}, 2023.

\bibitem[Pawlowski et~al.(2020)Pawlowski, Coelho~de Castro, and Glocker]{pawlowski2020deep}
Pawlowski, N., Coelho~de Castro, D., and Glocker, B.
\newblock Deep structural causal models for tractable counterfactual inference.
\newblock \emph{Advances in Neural Information Processing Systems}, 33:\penalty0 857--869, 2020.

\bibitem[Pearl(2000)]{Pearl2000}
Pearl, J.
\newblock \emph{Causality: Models, Reasoning and Inference}.
\newblock Cambridge University Press, 2000.

\bibitem[Pearl et~al.(2016)Pearl, Glymour, and Jewell]{Pearl2016Primer}
Pearl, J., Glymour, M., and Jewell, N.~P.
\newblock \emph{Causal inference in statistics: A primer}.
\newblock John Wiley Sons, 2016.

\bibitem[Powell(2020)]{powell2020quantile}
Powell, D.
\newblock Quantile treatment effects in the presence of covariates.
\newblock \emph{Review of Economics and Statistics}, 102\penalty0 (5):\penalty0 994--1005, 2020.

\bibitem[Ribeiro et~al.(2023)Ribeiro, Xia, Monteiro, Pawlowski, and Glocker]{ribeiro2023high}
Ribeiro, F. D.~S., Xia, T., Monteiro, M., Pawlowski, N., and Glocker, B.
\newblock High fidelity image counterfactuals with probabilistic causal models.
\newblock \emph{arXiv preprint arXiv:2306.15764}, 2023.

\bibitem[Sanchez \& Tsaftaris(2022)Sanchez and Tsaftaris]{sanchez2022diffusion}
Sanchez, P. and Tsaftaris, S.~A.
\newblock Diffusion causal models for counterfactual estimation.
\newblock \emph{arXiv preprint arXiv:2202.10166}, 2022.

\bibitem[Song et~al.(2020)Song, Meng, and Ermon]{song2020denoising}
Song, J., Meng, C., and Ermon, S.
\newblock Denoising diffusion implicit models.
\newblock \emph{arXiv preprint arXiv:2010.02502}, 2020.

\bibitem[Tagasovska \& Lopez-Paz(2019)Tagasovska and Lopez-Paz]{tagasovska2019single}
Tagasovska, N. and Lopez-Paz, D.
\newblock Single-model uncertainties for deep learning.
\newblock \emph{Advances in Neural Information Processing Systems}, 32, 2019.

\bibitem[Takeuchi et~al.(2006)Takeuchi, Le, Sears, Smola, et~al.]{takeuchi2006nonparametric}
Takeuchi, I., Le, Q., Sears, T., Smola, A., et~al.
\newblock Nonparametric quantile estimation.
\newblock 2006.

\bibitem[Takeuchi et~al.(2009)Takeuchi, Nomura, and Kanamori]{takeuchi2009nonparametric}
Takeuchi, I., Nomura, K., and Kanamori, T.
\newblock Nonparametric conditional density estimation using piecewise-linear solution path of kernel quantile regression.
\newblock \emph{Neural Computation}, 21\penalty0 (2):\penalty0 533--559, 2009.

\bibitem[Xia et~al.(2021)Xia, Lee, Bengio, and Bareinboim]{xia2021causal}
Xia, K., Lee, K.-Z., Bengio, Y., and Bareinboim, E.
\newblock The causal-neural connection: Expressiveness, learnability, and inference.
\newblock \emph{Advances in Neural Information Processing Systems}, 34:\penalty0 10823--10836, 2021.

\bibitem[Xia et~al.(2022)Xia, Pan, and Bareinboim]{xia2022neural}
Xia, K., Pan, Y., and Bareinboim, E.
\newblock Neural causal models for counterfactual identification and estimation.
\newblock \emph{arXiv preprint arXiv:2210.00035}, 2022.

\bibitem[Xie et~al.(2020)Xie, Cotton, and Zhu]{xie2020multiply}
Xie, Y., Cotton, C., and Zhu, Y.
\newblock Multiply robust estimation of causal quantile treatment effects.
\newblock \emph{Statistics in Medicine}, 39\penalty0 (28):\penalty0 4238--4251, 2020.

\bibitem[Yao et~al.(2018)Yao, Li, Li, Huai, Gao, and Zhang]{SITE2018}
Yao, L., Li, S., Li, Y., Huai, M., Gao, J., and Zhang, A.
\newblock Representation learning for treatment effect estimation from observational data.
\newblock In \emph{Neural Information Processing Systems}, 2018.

\bibitem[Yoon et~al.(2018)Yoon, Jordon, and Van Der~Schaar]{yoon2018ganite}
Yoon, J., Jordon, J., and Van Der~Schaar, M.
\newblock Ganite: Estimation of individualized treatment effects using generative adversarial nets.
\newblock In \emph{International conference on learning representations}, 2018.

\bibitem[Zhang et~al.(2015)Zhang, Wang, Zhang, and Sch{\"o}lkopf]{zhang2015estimation}
Zhang, K., Wang, Z., Zhang, J., and Sch{\"o}lkopf, B.
\newblock On estimation of functional causal models: general results and application to the post-nonlinear causal model.
\newblock \emph{ACM Transactions on Intelligent Systems and Technology (TIST)}, 7\penalty0 (2):\penalty0 1--22, 2015.

\bibitem[Zhou et~al.(2022)Zhou, Yao, Xu, Wang, and Zhu]{zhou2022cycle}
Zhou, G., Yao, L., Xu, X., Wang, C., and Zhu, L.
\newblock Cycle-balanced representation learning for counterfactual inference.
\newblock In \emph{Proceedings of the 2022 SIAM International Conference on Data Mining (SDM)}, pp.\  442--450. SIAM, 2022.

\bibitem[Zhou et~al.(2021{\natexlab{a}})Zhou, Carson~IV, and Carlson]{zhou2021estimating2}
Zhou, T., Carson~IV, W.~E., and Carlson, D.
\newblock Estimating potential outcome distributions with collaborating causal networks.
\newblock \emph{arXiv preprint arXiv:2110.01664}, 2021{\natexlab{a}}.

\bibitem[Zhou et~al.(2021{\natexlab{b}})Zhou, Li, Wu, and Carlson]{zhou2021estimating1}
Zhou, T., Li, Y., Wu, Y., and Carlson, D.
\newblock Estimating uncertainty intervals from collaborating networks.
\newblock \emph{The Journal of Machine Learning Research}, 22\penalty0 (1):\penalty0 11645--11691, 2021{\natexlab{b}}.

\end{thebibliography}
\bibliographystyle{icml2024}

\newpage
\appendix
\onecolumn
\section{Other Related Work}

Individual treatment effect (ITE) estimation is also closely related to the counterfactual inference problem. A major difference is that ITE focuses on the effect of treatment and the predictions of both factual and counterfactual outcomes on unseen samples. GANITE \citep{yoon2018ganite} first learns a counterfactual generator with GAN by matching the joint distribution of observed covariate and outcome variables, and then it generates a dataset by feeding different treatment values and random noises and learns an ITE generator to predict the factual and counterfactual outcomes. 
CFRNet \citep{johansson2016learning} models ITE as a domain adaptation problem where there is a distribution shift between effects under different treatment and match the marginal distributions of representations under different treatments in the representation space.  ABCEI \citep{du2021adversarial} proposes to use adversarial learning to balance the representations from treatment and control groups. CBRE \citep{zhou2022cycle} proposes to use cycle consistency to preserve the semantics of the representations from two groups.
 Based on GANITE, SCIGAN \citep{bica2020estimating} further proposes a hierarchical discriminator to learn the counterfactual generator when interventions are continuous, e.g., the dosage of the treatment. 
 SITE \citep{SITE2018} uses propensity score to select positive and negative pairs and proposes to minimize the middle point distance to preserve the relationships in the representation space. Based on SITE, CITE \citep{li2022contrastive} employs contrastive learning to preserve the relationships. BV-NICE \citep{lu2020reconsidering} models the generation process as a latent variable model where a confounder causes the treatment, covariate, and outcomes and addresses the covariate imbalance with adversarial training. 
The goal of \citep{zhou2021estimating1} is to estimate uncertainty intervals by learning two networks in an adversarial manner, where one is to estimate CDF and the other is to estimate the quantile. Later, \citep{zhou2021estimating2} extends \citep{zhou2021estimating1} to the ITE task. 
\citep{xie2020multiply,powell2020quantile} propose ways to estimate the quantile treatment effects unlike the average treatment effect. The quantile treatment effect is measured on all samples with same value of the quantile. On the contrary, our method learns different quantile for different individuals and use the quantile to represent the property of the conditional distribution.

\section{Proof of Theorem \ref{Theorem}}
\label{sec:proof_identifiability}

\begin{proof}[Proof]
Note $Y = f(X, Z, E)$ can be equivalently represented as $f_1(f_2(X,Z), g(E))$, and we denote $g(E)$ by $\tilde{E}$. 
We know that without further restrictions on the function class of $f$, the causal model $f$ and the probabilistic distribution $p(\tilde{E})$ are not identifiable \citep{zhang2015estimation}. Denote by $f^i$ and $p^i(\tilde{E})$ as one solution, and we will see that the counterfactual outcome actually does not depend on the index $i$; that is, it is independent of which $f^i$ and $P^i(\tilde{E}_{t+1})$ we choose.  Given observed evidence $(X=x, Y=y, Z=z)$, because $f^i$ is strictly monotonic in $\tilde{E}^i$, we can determine its value $\tilde{e}^i$, with $\tilde{e}^i = {f_{x,z}^i}^{-1}(y)$. Then, we can determine the value of the cumulative distribution function of $\tilde{E}^i$ at $\tilde{e}^i$, denoted by $\tau^i$.

Without loss of generality, we first show the case where $f^i$ is strictly increasing w.r.t. $\tilde{E}^i$. Because $f$ is strictly increasing in $\tilde{E}$ and $y=f^i(x,z,\tilde{e}^i)$, $y$ is the $\tau^i$-th quantile of $P(Y| X=x, Z=z)$. Then it is obvious that since $y$ and $P(Y |X=x, Z=z)$ are determined, the value of $\tau^i$ is independent of the index $i$, that is, it is identifiable. Thus, below, we will use $\tau$, instead of $\tau^i$. 

Since $g(E) \independent (X;Z)$, when doing interventions on $X$, the value $\tilde{e}^i$ will not change, as well as $e^i$. Hence, the counterfactual outcome $Y_{X=x'} |X=x, Y=y, Z=z$ can be calculated as $f^i(X=x',Z=z,E=e^i)$, and such equivalence can be directly derived following Pearl's three-step procedure. Because $\tilde{e}^i$ does not change after the intervention, the counterfactual outcome $Y_{X=x'} |X=x, Y=y, Z=z$ is the $\tau$-quantile of the conditional distribution $P(Y | X=x', Z=z)$. This quantile exists and it depends only on the conditional distribution $P(Y | X=x', Z=z)$, but not the chosen function $f^i$ and $P^i(\tilde{E})$. 

Therefore, the counterfactual outcome $Y_{X = x'} | X=x, Y=y, Z=z$ corresponds to the $\tau$-th quantile of the conditional distribution $P(Y|X=x', Z=z)$, where $Y=y$ represents the $\tau$-th quantile of $P(Y|X=x, Z=z)$.
\end{proof}

\section{Proof of Theorem \ref{theorem:bound}}
\label{sec:proof_of_bound}

Theorem 2 shows that the generalization error is bounded if we minimize the empirical loss, suggesting that our method learns a good quantile estimator using finite training samples. As a consequence (of this generalization bound and above identifiability theorem), our method is able to perform reliable counterfactual inference given finite training samples. 


In this section, we use $f$ to represent $\mu$. Further, for simplicity, we ignore $z$ in the $f$ function, which doesn't affect the generalization bound since the concatenation of $z$ and $x$ can be treated as a single input.  

We first give the following generalization error bound  derived by \citet{bartlett2002rademacher}.
\begin{Theorem}
Let $F$ be a hypothesis class mapping from $\mathcal{X}$ to $[0,b]$. Let $\{x_i\}_{i=1}^N$ be training samples with size $N$. Then, for any $\delta>0$, with probability at least $1-\delta$, the following holds for all $f\in F$:
\begin{flalign} \nonumber
\mathbb{E}_x[f(x)]\leq \frac{1}{N}\sum_{i=1}^{N}f(x_i)+ 2\mathfrak{R}(F)+b\sqrt{\frac{\log(1/\delta)}{2N}}.
\end{flalign}
\end{Theorem}

Inspired by the above theorem, we can derive a generalization error bound for counterfactual inference by nonlinear quantile regression.
\begin{Theorem}
\label{theorem:bound1}
Let $(\hat\tau,\hat{f}_{\hat\tau})\in(\mathfrak{T}, F)$ be the optimization solution. Let the loss function $l_\tau$ be upper bounded by $b$. Then, for any $\delta>0$, with probability at least $1-\delta$, we have
\begin{flalign} \nonumber
\mathbb{E}_{(x,y)}[l_{\hat\tau}(f^*(x) - \hat{f}(x))]\leq \frac{1}{N}\sum_{i=1}^{N}l_{\hat\tau}(f^*(x_i) - \hat{f}(x_i))+ 2\mathfrak{R}(\mathfrak{T}, F)+b\sqrt{\frac{\log(1/\delta)}{2N}},
\end{flalign}
where
\begin{flalign} \nonumber
\mathfrak{R}(\mathfrak{T}, F)=\mathbb{E}_{x,\sigma}\left[\sup_{\tau\in\mathfrak{T},f\in F}\frac{1}{N}\sum_{i=1}^N\sigma_il_{\tau}(f^*(x_i)-f(x_i))\right].
\end{flalign}
\end{Theorem}

Furthermore, we can derive the following result.
\begin{Theorem}
\label{theorem:bound2}
\begin{flalign} \nonumber
\mathfrak{R}(\mathfrak{T}, F)\leq 2\mathfrak{R}(F)+\frac{2b}{\sqrt{N}}.
\end{flalign}
\end{Theorem}

\textit{Proof.}
Let us rewrite $l_{\tau}(f^*(x)-f(x)) = \tau (f^*(x)-f(x))-1_{\{f^*(x)-f(x)<0\}}(f^*(x)-f(x))$, where $1_{\{A\}}$ is the indicator function. We have
\begin{flalign} \nonumber
\mathfrak{R}(\mathfrak{T}, F)&=\mathbb{E}_{x,\sigma}\left[\sup_{\tau\in\mathfrak{T},f\in F}\frac{1}{N}\sum_{i=1}^N\sigma_il_{\tau}(f^*(x_i)-f(x_i))\right]\nonumber\\
&=\mathbb{E}_{x,\sigma}\left[\sup_{\tau\in\mathfrak{T},f\in F}\frac{1}{N}\sum_{i=1}^N\sigma_i(\tau (f^*(x_i)-f(x_i))-1_{\{f^*(x_i)-f(x_i)<0\}}(f^*(x_i)-f(x_i)))\right]\nonumber\\
&\leq \mathbb{E}_{x,\sigma}\left[\sup_{\tau\in\mathfrak{T},f\in F}\tau\cdot\frac{1}{N}\sum_{i=1}^N\sigma_i (f^*(x_i)-f(x_i))\right]\nonumber\\
&+\mathbb{E}_{x,\sigma}\left[\sup_{f\in F}\frac{1}{N}\sum_{i=1}^N\sigma_i1_{\{f^*(x_i)-f(x_i)<0\}}(f^*(x_i)-f(x_i))\right]\nonumber\\
&= \mathbb{E}_{x,\sigma}\left[\sup_{f\in F}\frac{1}{N}\sum_{i=1}^N\sigma_i (f^*(x_i)-f(x_i))\right]\nonumber\\
&+\mathbb{E}_{x,\sigma}\left[\sup_{f\in F}\frac{1}{N}\sum_{i=1}^N\sigma_i1_{\{f^*(x_i)-f(x_i)<0\}}(f^*(x_i)-f(x_i))\right]\nonumber\\
&\leq 2\mathbb{E}_{x,\sigma}\left[\sup_{f\in F}\frac{1}{N}\sum_{i=1}^N\sigma_i (f^*(x_i)-f(x_i))\right]\nonumber\\
&= 2\mathbb{E}_{x,\sigma}\left[\sup_{f\in F}\frac{1}{N}\sum_{i=1}^N\sigma_i f(x_i)\right]+2\mathbb{E}_{x,\sigma}\left[\frac{1}{N}\sum_{i=1}^N\sigma_i f^*(x_i)\right]\nonumber\\
&\leq 2\mathfrak{R}(F)+\frac{2b}{\sqrt{N}}
\nonumber
\end{flalign}
where the second inequality holds because that $\mathfrak{R}(\phi\circ F)\leq L\mathfrak{R}(F)$ when $\phi: \mathbb{R}\rightarrow\mathbb{R}$ is an $L$-Lipschitz and that $1_{x<0}x$ is $1$-Lipschitz w.r.t. $x$. The last inequality holds because $\mathbb{E}_{x,\sigma}\left[\frac{1}{N}\sum_{i=1}^N\sigma_i f^*(x_i)\right]\leq\sqrt{\frac{b}{N}}$. Specifically,
\begin{flalign}
\mathbb{E}_{x,\sigma}\left[\frac{1}{N}\sum_{i=1}^N\sigma_i f^*(x_i)\right]&=\frac{1}{N}\mathbb{E}\left[\sum_{i=1}^N\sigma_i f^*(x_i)\right]\nonumber\\
&\leq \frac{1}{N}\left[\mathbb{E}\left[\left(\sum_{i=1}^N\sigma_i f^*(x_i)\right)^2\right]\right]^{1/2}\nonumber\\
&=\frac{1}{N}\left[\mathbb{E}\left[\sum_{i,j=1}^N\sigma_i\sigma_j f^*(x_i)f^*(x_j)\right]\right]^{1/2}\nonumber\\
&=\frac{1}{N}\left[\mathbb{E}\left[\sum_{i=1}^N (f^*(x_i))^2\right]\right]^{1/2}\nonumber\\
&\leq \frac{b}{\sqrt{N}}
\end{flalign}
where the first inequality follows by Jensen's inequality, the third equation holds because $\{\sigma_i\}_{i=1}^N$ are i.i.d. and $\mathbb{E}[\sigma_i\sigma_j] = \mathbb{E}[\sigma_i]\mathbb{E}[\sigma_j]=0$ when $i\neq j$ while $\mathbb{E}[\sigma_i\sigma_i] =1$.

Therefore, Theorem \ref{theorem:bound} can be derived by combining Theorem \ref{theorem:bound1} and \ref{theorem:bound2}.

\section{Datasets}

\textbf{Datasets}. Since our main result is based on the monotonicity assumption and the observations in counterfactual scenarios are lacking in real-world, we create following datasets.

\textbf{Rotation-MNIST}. We use MNIST images \cite{lecun2010mnist} as $Z$ and rotation angles as $X$. Then we randomly sample a noise value from $U[0,1]$ and adjust the  rotated images by adding values to the RGB channels. Therefore, the final pixel values are strictly monotonic w.r.t the noise value when conditioning on $X$ and $Z$. The training set consists of 60000 images while the testing set consists of 10000 images. The dimension of $Z$ is $3\times 32\times 32=3072$ and the dimension of $X$ is 1, which determines the rotation angles within [-45,45]. The dimension of $Y$ is also $3\times 32 \times 32$. In counterfactual inference, we are given a sample $X,Z,Y$, the question asks what had the images would be if the rotation angle is set to another value. In other words, we aim to rotate the digits in $Y$ while avoiding any other pixel value changes.

\textbf{Thick-Omniglot}. We use Omniglot images \cite{lake2019omniglot} as $Z$ and the thickness of characters as $X$. Then we randomly sample a noise value from $U[0,1]$ and adjust the background darkness by multiplying the thicken characters pixel values. Therefore, the final pixel values are strictly monotonic w.r.t the noise value when conditioning on $X$ and $Z$. The number of images is 19280 and We split the dataset as 80/20. We also resize the images to 32$\times$32. In this dataset, our goal is to adjust the thickness by the new value of $X$ while preserving the brightness in $Y$. 

\textbf{Cont-Dose}. We generate this dataset by setting $X$ as the treatment and $Z$ as the age of patient and the outcome $Y$ denotes the effect. $X$ takes value from [0,2] with space 0.1 uniformly. The numbers of training samples is 800 and the number of testing samples is 200.

\textbf{Dis-Dose}. We generate this dataset by setting $X$ as the treatment and $Z$ as the age of patient and the outcome $Y$ denotes the effect. $X$ takes value from $\{0,1\}$. The numbers of training samples is 800 and the number of testing samples is 200.

We also use the semi-simulated dataset IHDP \cite{Hill2011}. It contains 100 splits. Each split consists of 675 training and 92 testing samples. The dimension of $Z$ is 25 and the dimension of $X$ is 1 and the dimension of output $Y$ is 1.

\section{Implementation}
\textbf{All codes are provided in the supplementary materials. Readers may refer to the code for more implementation details.}

We use betty \citep{choe2023betty} as our bi-level optimization library. 
Then we define the lower level loss as the pin-ball loss with the quantile passed from the upper level. After optimizing the lower level net $g$ with 30 iterations, we fix the network $g$ and optimize the upper-level network $h$ on the regression loss (We adopt MSE loss). 

For toy datasets and cont-Dose dataset, we build the upper-level network as 3 layers and 1 output linear layer. The network $h$ takes as input the factual observations $X$, $Z$ and $Y$. Then we concatenate the three variables  and feed into the model. The first layer is a linear layer transforming the input into a hidden-dimension (200) and apply a SiLU activation function. Then we use 2 residual blocks (with layernorm) and finally feed the output into the linear layer with output size as 1. The final activation is Sigmoid since we need to use $h$ to represent quantile, which is within range [0,1]. As for the lower-level network $g$,  it is mainly trained to get a regression at quantile $\tau$, which is the output of network $h$. We first project the input $X,Z,\tau$ into same dimensions respectively, then we feed into a linear layer to match the dimension of $Y$. 

For Dis-Dose and IHDP, $X$ is binary. Therefore, we use two networks with same architecture as $h$. We feed the inputs $X,Z,Y$ into different sub-network according to the value $X$. We find that a shallower network is better, therefore, we do not use residual block here.  For the network $h$, we use 3-layers (one input layer and 2 residual blocks).

For image transformation datasets, we use the convolutional neural network to parameterize $h$ and $g$. The hidden dimension of $h$ is 32 while the hidden dim of $g$ is 128. We downsample the images with Conv(4,2,1) and final conv(4,1,0) to get the quantile estimations. In network $g$, we use a symmetric network where the upsamplinmg network mimic the downsampling process with ConvTranspose(4,2,1).

We use $n_i$ interested samples in training network $g$. For eacch interested sample, we have one quantile $\tau$. Then we train the network $g$ with $n_t$ samples for each $\tau$. We set $n_i=256, n_t=64$ for most cases. As for image datasets, we have to lower the values to reduce memory. So we use $n_i=128, n_t=32$. In the upper-level problem, we use 64 (128) samples to get the reconstruction loss. We use Adam optimizer with lr=2e-3 for images while 1e-3 for the rest of them.

 We choose the hyper-parameters based on the values on some toy datasets since we can always generate them. Then we apply the best hyper-parameters to the formal dataset, Dis-Dose, Cont-Dose, IHDP,  Thick-Ominiglot and Rotation-MNIST. During training, we use the reconstruction error on the training dataset as metric to select the models.

\section{Confounding}

In the main paper, we considered a simple case where $Y=X+Z+E$, where $Z \sim U[0,1], X \sim U[0,1]$ and $e\sim N(0,1)$. We generate the latent confounder as $C\sim U[-0.5, 0.5]$. Then we add the confounder to the its children in dfferent scenarios. For example, when we have $Z\leftarrow C \rightarrow Y$, we add $Z=Z+C, Y=Y+C$. According  to the table in the main paper, the performances drop on the scenario $X \leftarrow C \rightarrow Y$ in this case, which is kindly expected since we do not assume the existence of latent confounder and $C$ influences $X$ and $Y$ now. (the code of data generation and training is provided in the supp.)

\section{More Results on Image Transformation Datset}

\begin{figure}
    \centering
 \begin{tabular}{cccccccc}
    Z &E= 0.1 & 0.3 & 0.5 &  0.7 & 0.9 \\ \hline
    &$\hat{\tau}=$  0.1075 & 0.3088 & 0.5094 & 0.7032 &0.9019\\ \hline
      \includegraphics[scale=1]{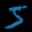}&
         \includegraphics[scale=1]{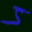}&
           \includegraphics[scale=1]{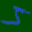}&
           \includegraphics[scale=1]{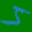}&
           \includegraphics[scale=1]{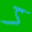}&
           \includegraphics[scale=1]{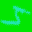}&
      \\ \hline
      & $\hat{\tau}$=  0.1034 & 0.2945 & 0.5002 & 0.6984 & 0.8972\\ \hline
         \includegraphics[scale=1]{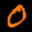}&
         \includegraphics[scale=1]{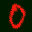}&
           \includegraphics[scale=1]{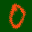}&
           \includegraphics[scale=1]{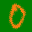}&
           \includegraphics[scale=1]{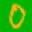}&
           \includegraphics[scale=1]{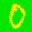}&
       \\ \hline
        &$\hat{\tau}$=  0.0979 & 0.3069 & 0.5019 & 0.7060 & 0.9080\\ \hline
        \includegraphics[scale=1]{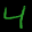}&
         \includegraphics[scale=1]{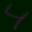}&
           \includegraphics[scale=1]{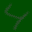}&
           \includegraphics[scale=1]{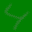}&
           \includegraphics[scale=1]{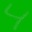}&
           \includegraphics[scale=1]{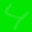}&
        \\ \hline
        &$\hat{\tau}$=0.1045 & 0.3026 & 0.4977 & 0.7003 & 0.8997\\ \hline
          \includegraphics[scale=1]{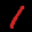}&
         \includegraphics[scale=1]{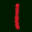}&
           \includegraphics[scale=1]{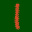}&
           \includegraphics[scale=1]{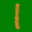}&
           \includegraphics[scale=1]{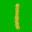}&
           \includegraphics[scale=1]{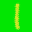}&
           \\ \hline
          & $\hat{\tau}$=  0.1057 & 0.3050 &  0.5024 & 0.7027 & 0.8970\\ \hline
  \includegraphics[scale=1]{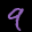}&
         \includegraphics[scale=1]{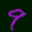}&
           \includegraphics[scale=1]{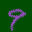}&
           \includegraphics[scale=1]{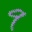}&
           \includegraphics[scale=1]{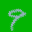}&
           \includegraphics[scale=1]{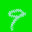}&
           \\ \hline
  & $\hat{\tau}$=  0.1059 & 0.3032 & 0.5013 & 0.6956 & 0.8972\\ \hline
   \includegraphics[scale=1]{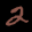}&
         \includegraphics[scale=1]{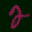}&
           \includegraphics[scale=1]{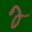}&
           \includegraphics[scale=1]{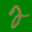}&
           \includegraphics[scale=1]{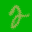}&
           \includegraphics[scale=1]{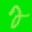}&
           \\ \hline
            & $\hat{\tau}$= 0.1059 & 0.3021 & 0.5012 & 0.7022 & 0.8967\\ \hline
         \includegraphics[scale=1]{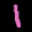}&
         \includegraphics[scale=1]{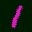}&
           \includegraphics[scale=1]{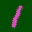}&
           \includegraphics[scale=1]{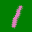}&
           \includegraphics[scale=1]{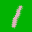}&
           \includegraphics[scale=1]{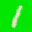}&
          \\ \hline
  & $\hat{\tau}$= 0.1051 & 0.2991 & 0.5038 & 0.7017 & 0.8989\\ \hline
   \includegraphics[scale=1]{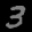}&
         \includegraphics[scale=1]{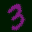}&
           \includegraphics[scale=1]{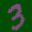}&
           \includegraphics[scale=1]{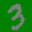}&
           \includegraphics[scale=1]{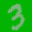}&
           \includegraphics[scale=1]{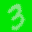}&
           \\ \hline
   & $\hat{\tau}$=  0.1007 & 0.2998 & 0.4979 & 0.7030 & 0.9027\\ \hline
     \includegraphics[scale=1]{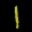}&
         \includegraphics[scale=1]{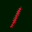}&
           \includegraphics[scale=1]{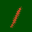}&
           \includegraphics[scale=1]{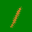}&
           \includegraphics[scale=1]{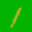}&
           \includegraphics[scale=1]{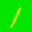}&
           \\ \hline
   & $\hat{\tau}$=  0.1015 & 0.3015 & 0.4989 & 0.6986 & 0.8981\\ \hline
      \includegraphics[scale=1]{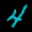}&
         \includegraphics[scale=1]{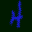}&
           \includegraphics[scale=1]{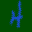}&
           \includegraphics[scale=1]{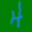}&
           \includegraphics[scale=1]{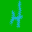}&
           \includegraphics[scale=1]{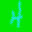}&
           \\ \hline
    
 \end{tabular}
    \caption{Examples of learned quantiles on Rotation-MNIST dataset. The learned quantiles $\hat{\tau}$ are very close to the true quantiles, i.e., 0.1, 0.3,0.5,0.7 and 0.9. }
    \label{fig:append_quantiles_mnist}
\end{figure}

\begin{figure}
    \centering
 \begin{tabular}{cccccccc}
    Z &E= 0.1 & 0.3 & 0.5 &  0.7 & 0.9 \\ \hline
    &$\hat{\tau}=$  0.0908 & 0.3034 & 0.5011 & 0.7062 &0.8981\\ \hline
      \includegraphics[scale=1]{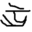}&
         \includegraphics[scale=1]{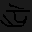}&
           \includegraphics[scale=1]{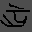}&
            \includegraphics[scale=1]{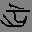}&
           \includegraphics[scale=1]{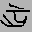}&
           \includegraphics[scale=1]{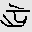}
      \\ \hline
      & $\hat{\tau}$=  0.0934 & 0.3011 & 0.5014 & 0.7082 & 0.8929\\ \hline
       \includegraphics[scale=1]{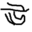}&
        \includegraphics[scale=1]{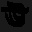}&
           \includegraphics[scale=1]{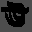}&
             \includegraphics[scale=1]{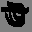}&
           \includegraphics[scale=1]{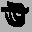}&
           \includegraphics[scale=1]{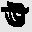}
       \\ \hline
        &$\hat{\tau}$=  0.0934 & 0.3055 & 0.5040 & 0.7089 & 0.8982\\ \hline
        \includegraphics[scale=1]{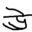}& 
         \includegraphics[scale=1]{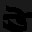}&
           \includegraphics[scale=1]{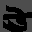}&
             \includegraphics[scale=1]{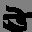}&
           \includegraphics[scale=1]{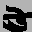}&
           \includegraphics[scale=1]{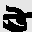}
        \\ \hline
        &$\hat{\tau}$=0.0932 & 0.2901 & 0.4612 & 0.6431 & 0.8239\\ \hline
         \includegraphics[scale=1]{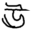}&
          \includegraphics[scale=1]{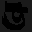}&
           \includegraphics[scale=1]{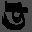}&
             \includegraphics[scale=1]{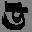}&
           \includegraphics[scale=1]{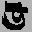}&
           \includegraphics[scale=1]{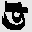}
           \\ \hline
          & $\hat{\tau}$=  0.1028 & 0.2972 &  0.4956 & 0.7003 & 0.8961\\ \hline
 \includegraphics[scale=1]{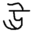}& 
  \includegraphics[scale=1]{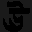}&
           \includegraphics[scale=1]{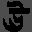}&
             \includegraphics[scale=1]{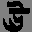}&
           \includegraphics[scale=1]{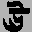}&
           \includegraphics[scale=1]{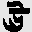}
           \\ \hline
  & $\hat{\tau}$=  0.0863 & 0.3079 & 0.4936 & 0.6877 & 0.8650\\ \hline
  \includegraphics[scale=1]{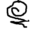}&  \includegraphics[scale=1]{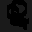}&
           \includegraphics[scale=1]{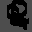}&
             \includegraphics[scale=1]{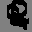}&
           \includegraphics[scale=1]{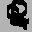}&
           \includegraphics[scale=1]{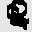}
           \\ \hline
  & $\hat{\tau}$= 0.1076 & 0.3023 & 0.4953 & 0.6999 & 0.8960\\ \hline
   \includegraphics[scale=1]{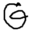}&
    \includegraphics[scale=1]{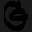}&
           \includegraphics[scale=1]{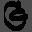}&
             \includegraphics[scale=1]{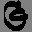}&
           \includegraphics[scale=1]{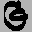}&
           \includegraphics[scale=1]{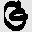}
           \\ \hline
   & $\hat{\tau}$=  0.0938 & 0.2995 & 0.4868 & 0.6837 & 0.8648\\ \hline
    \includegraphics[scale=1]{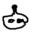}& 
     \includegraphics[scale=1]{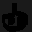}&
           \includegraphics[scale=1]{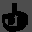}&
             \includegraphics[scale=1]{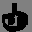}&
           \includegraphics[scale=1]{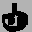}&
           \includegraphics[scale=1]{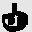}
           \\ \hline
   & $\hat{\tau}$=  0.0991 & 0.3050 & 0.4993 & 0.7036 & 0.8947\\ \hline
     \includegraphics[scale=1]{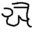}&
      \includegraphics[scale=1]{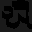}&
           \includegraphics[scale=1]{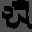}&
             \includegraphics[scale=1]{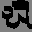}&
           \includegraphics[scale=1]{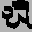}&
           \includegraphics[scale=1]{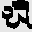}
           \\ \hline
     & $\hat{\tau}$= 0.0946 & 0.2982 & 0.5014 & 0.7107 & 0.8964\\ \hline
          \includegraphics[scale=1]{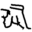}&
           \includegraphics[scale=1]{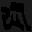}&
           \includegraphics[scale=1]{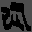}&
             \includegraphics[scale=1]{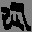}&
           \includegraphics[scale=1]{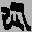}&
           \includegraphics[scale=1]{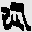}
          \\ \hline
 \end{tabular}
    \caption{Examples of learned quantiles on Thick-Omniglot dataset. The learned quantiles $\hat{\tau}$ are very close to the true quantiles, i.e., 0.1, 0.3,0.5,0.7 and 0.9. Each row displays the images after adjusting thickness by $X$ and the brightness with noise value. The noise determines the brightness of the images. From the $Z$ and the output image, our method is not affected by the changing thickness between $Z$ and $Y$ and able to capture the brightness changes. The accurate estimation shows the effectiveness of our approach even when $Z$ and $Y$ are high-dimensional.}
    \label{fig:append_quantiles_omni}
\end{figure}

\textbf{Learned Quantiles}.
Since we sample the noise values uniformly from [0,1]. The target quantile $P(Y\leq y|X=x, Z=z)$ is the CDF of the noise. In other words, if the noise is $a$, the target quantile should be $a$.  
We present the examples of Thick-Omniglot in Fig. \ref{fig:append_quantiles_omni}. Our estimated quantiles are not affected by the thickness changes between the input $Z$ and the output $Y$. The The learned quantiles are very close to the ground truth, further demonstrating the effectiveness of our method. We also provide examples of Rotation-MNIST in Fig. \ref{fig:append_quantiles_mnist}. Our method is able recover the quantiles of the true noises accurately.

\textbf{Counterfactual Prediction Results} We present more visual examples of Rotation-MNIST in Fig. \ref{fig:append_mnist1} and \ref{fig: append_mnist2}. We can see that the CFQP \cite{de2022deep} learns the rotation when changing the value of $X$. However, the rotations results is unsatisfactory. In particular, the background color of the rotated digits are different from the ground truth $Y_{X=x^\prime}$ and the $Y$. 
This is unwanted because $X$ only affects the rotation.
It indicates that CFQP fails to capture the uniqueness (noise) of the sample and leads to unnecessary changes. As for BGM \cite{nasr2023counterfactual2}, the images are almost identical to the $Y$. The reason is that the BGM employs conditional spline flow to mimic the strictly monotonic generation process. However, it is known that the expressive power of flow, especially such strictly monotonic flow, are limited compared to the unconstrained neural networks. It fails to transform the high-dimensional input $Y$ into gaussian noise and therfore, fails to utilize the conditions ($X$ and $Z$). As a consequence, when we change the value of $X$, BGM ignores the condition and generates almost identical images to the input $Y$. DCM \cite{chao2023interventional} models the generation process with conditional diffusion model and recovers the noise by inverting the diffusion process. However, since the true noise is usually not Gaussian, the transformed noise may not be meaningful to counterfactual inference. Therefore, the counterfactual predictions are not accurate. In contrast to above baseline methods, our approach learns the correct rotation as well as preserving the correct color in the factual observation $Y$.

\begin{figure}[ht]
    \centering
   \begin{tabular}{ccccccc}
     Z & Y & CFQP & BGM & DCM & Ours & Truth  \\
       \includegraphics[scale=1]{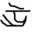} &
\includegraphics[scale=1]{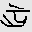} & 
\includegraphics[scale=1]{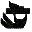} &
\includegraphics[scale=1]{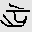} &
\includegraphics[scale=1]{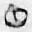} & 
\includegraphics[scale=1]{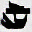} &
\includegraphics[scale=1]{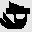}\\
       \includegraphics[scale=1]{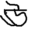} &
\includegraphics[scale=1]{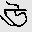} & 
\includegraphics[scale=1]{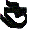} &
\includegraphics[scale=1]{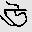} &
\includegraphics[scale=1]{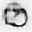} & 
\includegraphics[scale=1]{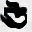} &
\includegraphics[scale=1]{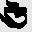}\\
       \includegraphics[scale=1]{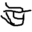} &
\includegraphics[scale=1]{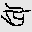} & 
\includegraphics[scale=1]{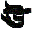} &
\includegraphics[scale=1]{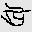} &
\includegraphics[scale=1]{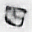} & 
\includegraphics[scale=1]{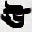} &
\includegraphics[scale=1]{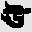}\\
       \includegraphics[scale=1]{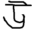} &
\includegraphics[scale=1]{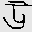} & 
\includegraphics[scale=1]{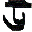} &
\includegraphics[scale=1]{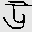} &
\includegraphics[scale=1]{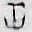} & 
\includegraphics[scale=1]{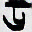} &
\includegraphics[scale=1]{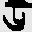}\\
       \includegraphics[scale=1]{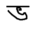} &
\includegraphics[scale=1]{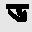} & 
\includegraphics[scale=1]{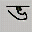} &
\includegraphics[scale=1]{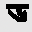} &
\includegraphics[scale=1]{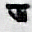} & 
\includegraphics[scale=1]{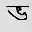} &
\includegraphics[scale=1]{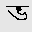}\\
       \includegraphics[scale=1]{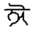} &
\includegraphics[scale=1]{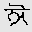} & 
\includegraphics[scale=1]{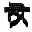} &
\includegraphics[scale=1]{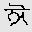} &
\includegraphics[scale=1]{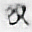} & 
\includegraphics[scale=1]{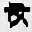} &
\includegraphics[scale=1]{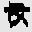}\\
       \includegraphics[scale=1]{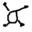} &
\includegraphics[scale=1]{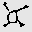} & 
\includegraphics[scale=1]{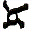} &
\includegraphics[scale=1]{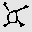} &
\includegraphics[scale=1]{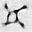} & 
\includegraphics[scale=1]{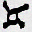} &
\includegraphics[scale=1]{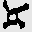}\\
       \includegraphics[scale=1]{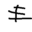} &
\includegraphics[scale=1]{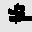} & 
\includegraphics[scale=1]{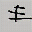} &
\includegraphics[scale=1]{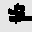} &
\includegraphics[scale=1]{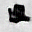} & 
\includegraphics[scale=1]{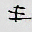} &
\includegraphics[scale=1]{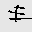}\\
       \includegraphics[scale=1]{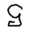} &
\includegraphics[scale=1]{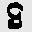} & 
\includegraphics[scale=1]{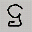} &
\includegraphics[scale=1]{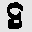} &
\includegraphics[scale=1]{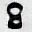} & 
\includegraphics[scale=1]{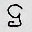} &
\includegraphics[scale=1]{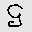}\\
       \includegraphics[scale=1]{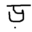} &
\includegraphics[scale=1]{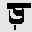} & 
\includegraphics[scale=1]{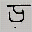} &
\includegraphics[scale=1]{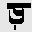} &
\includegraphics[scale=1]{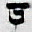} & 
\includegraphics[scale=1]{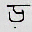} &
\includegraphics[scale=1]{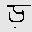}\\
   \end{tabular}
    \caption{Examples of counterfactual predictions on Thick-Omniglot dataset.}
    \label{fig:append_omni}
\end{figure}

We also present examples of Thick-Omniglot in Fig. \ref{fig:append_omni}. The noises values determines the darkness of the image. We obvserve that CFQP \cite{de2022deep} fails to preserve the darkness of the image $Y$. Sometimes the predictions are darker or much brighter. DCM \cite{chao2023interventional} struggles to adjust the thickness of the images. In contrast, our method learns to preserve the brightness in the factual observation $Y$ while changing the thickness of the digits.

\begin{table}[ht]
    \centering
    \begin{tabular}{ccccccc}
       Z & Y & CFQP & BGM & DCM & Ours & Truth  \\
\includegraphics[scale=1]{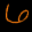} &
\includegraphics[scale=1]{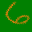} & 
\includegraphics[scale=1]{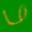} &
\includegraphics[scale=1]{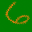} &
\includegraphics[scale=1]{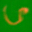} & 
\includegraphics[scale=1]{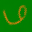} &
\includegraphics[scale=1]{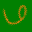}\\
\includegraphics[scale=1]{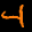} &
\includegraphics[scale=1]{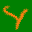} & 
\includegraphics[scale=1]{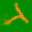} &
\includegraphics[scale=1]{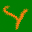} &
\includegraphics[scale=1]{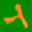} & 
\includegraphics[scale=1]{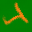} &
\includegraphics[scale=1]{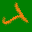}\\
\includegraphics[scale=1]{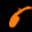} &
\includegraphics[scale=1]{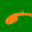} & 
\includegraphics[scale=1]{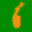} &
\includegraphics[scale=1]{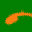} &
\includegraphics[scale=1]{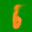} & 
\includegraphics[scale=1]{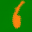} &
\includegraphics[scale=1]{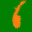}\\
\includegraphics[scale=1]{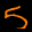} &
\includegraphics[scale=1]{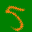} & 
\includegraphics[scale=1]{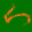} &
\includegraphics[scale=1]{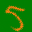} &
\includegraphics[scale=1]{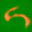} & 
\includegraphics[scale=1]{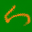} &
\includegraphics[scale=1]{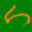}\\
\includegraphics[scale=1]{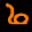} &
\includegraphics[scale=1]{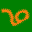} & 
\includegraphics[scale=1]{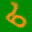} &
\includegraphics[scale=1]{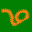} &
\includegraphics[scale=1]{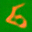} & 
\includegraphics[scale=1]{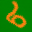} &
\includegraphics[scale=1]{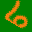}\\
\includegraphics[scale=1]{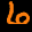} &
\includegraphics[scale=1]{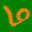} & 
\includegraphics[scale=1]{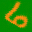} &
\includegraphics[scale=1]{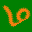} &
\includegraphics[scale=1]{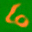} & 
\includegraphics[scale=1]{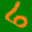} &
\includegraphics[scale=1]{supp_figs/mnist/4841_cf_color1.png}\\
\includegraphics[scale=1]{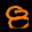} &
\includegraphics[scale=1]{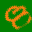} & 
\includegraphics[scale=1]{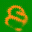} &
\includegraphics[scale=1]{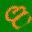} &
\includegraphics[scale=1]{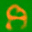} & 
\includegraphics[scale=1]{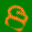} &
\includegraphics[scale=1]{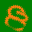}\\
\includegraphics[scale=1]{supp_figs/mnist/4952_cov_color1.png} &
\includegraphics[scale=1]{supp_figs/mnist/4952_out_color1.png} & 
\includegraphics[scale=1]{supp_figs/mnist/4952_cfqp_color1.png} &
\includegraphics[scale=1]{supp_figs/mnist/bgm_4952_color1.png} &
\includegraphics[scale=1]{supp_figs/mnist/dcm_4952_color1.png} & 
\includegraphics[scale=1]{supp_figs/mnist/pred_4952_color1.png} &
\includegraphics[scale=1]{supp_figs/mnist/4952_cf_color1.png}\\
\includegraphics[scale=1]{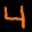} &
\includegraphics[scale=1]{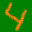} & 
\includegraphics[scale=1]{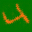} &
\includegraphics[scale=1]{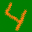} &
\includegraphics[scale=1]{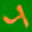} & 
\includegraphics[scale=1]{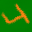} &
\includegraphics[scale=1]{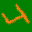}\\
\includegraphics[scale=1]{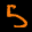} &
\includegraphics[scale=1]{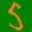} & 
\includegraphics[scale=1]{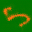} &
\includegraphics[scale=1]{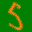} &
\includegraphics[scale=1]{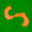} & 
\includegraphics[scale=1]{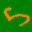} &
\includegraphics[scale=1]{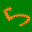}\\
    \end{tabular}
    \caption{Examples of  counterfactual predictions on Rotation-MNIST.  }
    \label{fig:append_mnist1}
\end{table}

\begin{table}[ht]
    \centering
    \begin{tabular}{ccccccc}
       Z & Y & CFQP & BGM & DCM & Ours & Truth  \\
\includegraphics[scale=1]{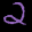} &
\includegraphics[scale=1]{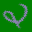} & 
\includegraphics[scale=1]{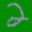} &
\includegraphics[scale=1]{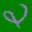} &
\includegraphics[scale=1]{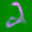} & 
\includegraphics[scale=1]{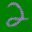} &
\includegraphics[scale=1]{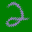}\\
\includegraphics[scale=1]{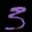} &
\includegraphics[scale=1]{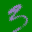} & 
\includegraphics[scale=1]{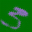} &
\includegraphics[scale=1]{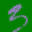} &
\includegraphics[scale=1]{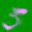} & 
\includegraphics[scale=1]{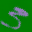} &
\includegraphics[scale=1]{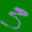}\\
\includegraphics[scale=1]{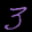} &
\includegraphics[scale=1]{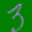} & 
\includegraphics[scale=1]{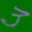} &
\includegraphics[scale=1]{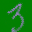} &
\includegraphics[scale=1]{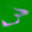} & 
\includegraphics[scale=1]{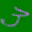} &
\includegraphics[scale=1]{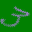}\\
\includegraphics[scale=1]{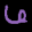} &
\includegraphics[scale=1]{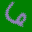} & 
\includegraphics[scale=1]{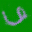} &
\includegraphics[scale=1]{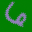} &
\includegraphics[scale=1]{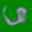} & 
\includegraphics[scale=1]{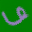} &
\includegraphics[scale=1]{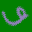}\\
\includegraphics[scale=1]{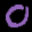} &
\includegraphics[scale=1]{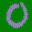} & 
\includegraphics[scale=1]{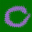} &
\includegraphics[scale=1]{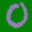} &
\includegraphics[scale=1]{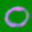} & 
\includegraphics[scale=1]{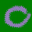} &
\includegraphics[scale=1]{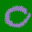}\\
\includegraphics[scale=1]{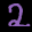} &
\includegraphics[scale=1]{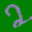} & 
\includegraphics[scale=1]{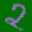} &
\includegraphics[scale=1]{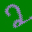} &
\includegraphics[scale=1]{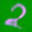} & 
\includegraphics[scale=1]{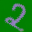} &
\includegraphics[scale=1]{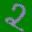}\\
\includegraphics[scale=1]{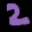} &
\includegraphics[scale=1]{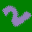} & 
\includegraphics[scale=1]{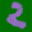} &
\includegraphics[scale=1]{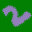} &
\includegraphics[scale=1]{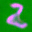} & 
\includegraphics[scale=1]{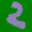} &
\includegraphics[scale=1]{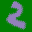}\\
\includegraphics[scale=1]{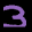} &
\includegraphics[scale=1]{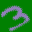} & 
\includegraphics[scale=1]{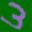} &
\includegraphics[scale=1]{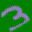} &
\includegraphics[scale=1]{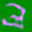} & 
\includegraphics[scale=1]{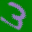} &
\includegraphics[scale=1]{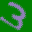}\\
\includegraphics[scale=1]{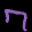} &
\includegraphics[scale=1]{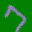} & 
\includegraphics[scale=1]{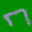} &
\includegraphics[scale=1]{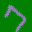} &
\includegraphics[scale=1]{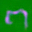} & 
\includegraphics[scale=1]{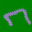} &
\includegraphics[scale=1]{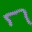}\\
\includegraphics[scale=1]{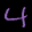} &
\includegraphics[scale=1]{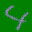} & 
\includegraphics[scale=1]{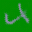} &
\includegraphics[scale=1]{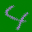} &
\includegraphics[scale=1]{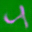} & 
\includegraphics[scale=1]{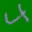} &
\includegraphics[scale=1]{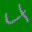}\\
    \end{tabular}
    \caption{Examples of  counterfactual predictions on Rotation-MNIST. CFQP fails to preserve the brightness of image $Y$ in their counterfactual predictions. BGM fails to transform the image $Y$ given the limited expressive power of the used spline flow. DCM genrates unrealistic digits. }
    \label{fig: append_mnist2}
\end{table}

\end{document}